\documentclass{article}

\usepackage{PRIMEarxiv}

\usepackage[utf8]{inputenc} 
\usepackage[T1]{fontenc}    
\usepackage{hyperref}       
\usepackage{url}            
\usepackage{booktabs}       
\usepackage{amsfonts}       
\usepackage{nicefrac}       
\usepackage{microtype}      
\usepackage{lipsum}
\usepackage{fancyhdr}       
\usepackage{graphicx}       
\graphicspath{{media/}}     

\usepackage{times}
\usepackage{soul}
\usepackage{amsmath}
\usepackage{amsthm}
\usepackage{booktabs}
\usepackage{algorithm}
\usepackage{algorithmic}
\usepackage{enumerate}
\usepackage{subfigure}
\usepackage{multirow}
\usepackage{enumitem}
\usepackage{xspace}
\usepackage{multicol}
\usepackage{diagbox}
\usepackage{url}
\usepackage{algorithm}
\usepackage{algorithmic}
\usepackage{array}
\usepackage{xspace}
\usepackage{booktabs}      
\usepackage{amsfonts}      
\usepackage{nicefrac}       
\usepackage{microtype}      
\usepackage{multirow}
\usepackage{enumerate}
\usepackage{makecell}
\usepackage{epstopdf}
\usepackage{graphics}
\usepackage{graphicx}
\usepackage{xcolor}
\usepackage{enumitem}
\usepackage{bbm}
\usepackage{amssymb, latexsym, amsmath,bm}
\usepackage{multicol}
\usepackage{diagbox}
\newtheorem{definition}{Definition} 
\newtheorem{property}{Property}
\newtheorem{theorem}{Theorem} 
\newtheorem{task}{Task}

\newcommand{\mname}{$\texttt{CRUCIAL}$\xspace}
\newcommand{\mnamee}{$\texttt{CRUCIAL(SIN)}$\xspace}
\newcommand{\mnameee}{$\texttt{CRUCIAL(ADP)}$\xspace}

\title{Curricular and Cyclical Loss\\ for Time Series Learning Strategy}

\author{
  Chenxi Sun \\
  Key Laboratory of Machine Perception\\
  (Ministry of Education), Peking University\\
  Beijing, China\\
  National Key Laboratory of General Artificial Intelligence\\
  Beijing, China\\
  School of Artificial Intelligence,\\ Peking University\\
  Beijing, China\\
  \texttt{sun\_chenxi@pku.edu.cn} \\
\And
  Hongyan Li* \\
  Key Laboratory of Machine Perception\\
  (Ministry of Education), Peking University\\
  Beijing, China\\
  National Key Laboratory of General Artificial Intelligence\\
  Beijing, China\\
  School of Artificial Intelligence,\\ Peking University\\
  Beijing, China\\
   \And
  Moxian Song \\
  Key Laboratory of Machine Perception\\
  (Ministry of Education), Peking University\\
  Beijing, China\\
  School of Artificial Intelligence,\\ Peking University\\
  Beijing, China\\
    \And
  Derun Cai \\
  Key Laboratory of Machine Perception\\
  (Ministry of Education), Peking University\\
  Beijing, China\\
  School of Artificial Intelligence,\\ Peking University\\
  Beijing, China\\
    \And
  Shenda Hong* \\
  National Institute of Health Data Science,\\Peking University\\
  Beijing, China\\
  Institute of Medical Technology, \\Health Science Center of Peking University\\
  Beijing, China\\
}

\begin{document}
\maketitle

\begin{abstract}
Time series widely exists in real-world applications and many deep learning models have performed well on it. Current research has shown the importance of learning strategy for models, suggesting that the benefit is the order and size of learning samples. However, no effective strategy has been proposed for time series due to its abstract and dynamic construction. Meanwhile, the existing one-shot tasks and continuous tasks for time series necessitate distinct learning processes and mechanisms. No all-purpose approach has been suggested. In this work, we propose a novel Curricular and CyclicaL loss (\mname) to learn time series for the first time. It is model- and task-agnostic and can be plugged on top of the original loss with no extra procedure. \mname has two characteristics: It can arrange an easy-to-hard learning order by dynamically determining the sample contribution and modulating the loss amplitude; It can manage a cyclically changed dataset and achieve an adaptive cycle by correlating the loss distribution and the selection probability. We prove that compared with monotonous size, cyclical size can reduce expected error. Experiments on 3 kinds of tasks and 5 real-world datasets show the benefits of \mname for most deep learning models when learning time series. 
\end{abstract}


\section{Introduction}

\begin{figure*}[t]
\centering
\includegraphics[width=0.92\linewidth]{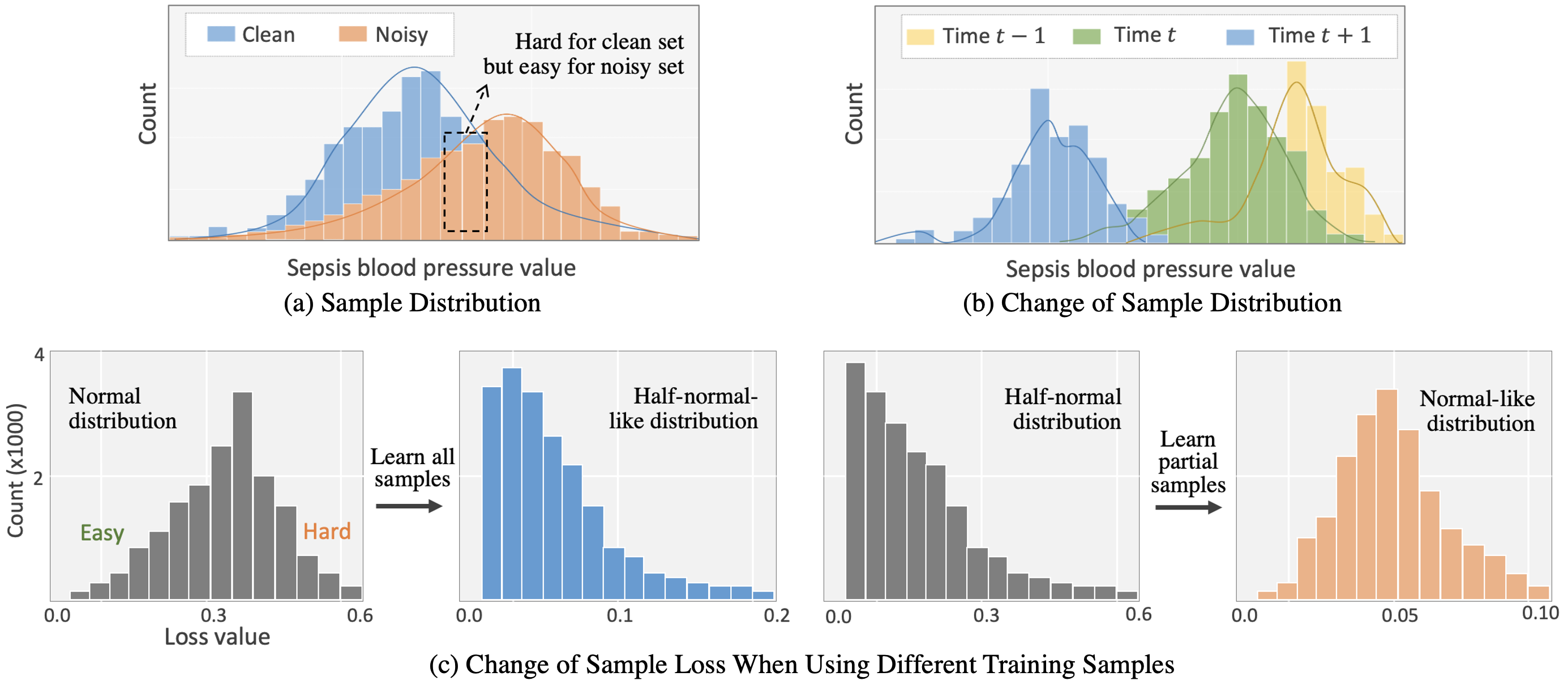} 
\caption{Observations When Deep Learning Models Learning Time Series}
\label{fig:observation}
\end{figure*}

At present, deep learning models have achieved good results in time series data, which has diverse sequence and shape features \cite{9210118,9721108} and widely exists in real-world applications such as medical and meteorological \cite{Sun_2022}. Many studies have shown that a meaningful learning mechanism can improve the model accuracy and generalization power \cite{9416768,DBLP:conf/cikm/SunSCZH022}. For example, an easy-to-hard curriculum can get better results than random shuffling on image data \cite{DBLP:conf/icml/HacohenW19}. 

However, the time series is too abstract to define its difficulty for curriculum in advance \cite{DBLP:journals/corr/abs-2010-12493}. As shown in Figure \ref{fig:observation} (a), a non-stationary time series could be a hard sample or easy sample under different noise and irregularity ratios \cite{DBLP:conf/icml/ArazoOAOM19,DBLP:conf/ijcai/SunHSCSC021}. Defining the learning order of time series data requires an extra program. It is often a separate process, does not adapt to models, and requires additional cost. At present, the learning mechanism for deep learning models is often defined as specific to datasets and tasks. A general, task-agnostic, and easy-to-use method is unaddressed. 

Meanwhile, the size of the dataset learned by the model in each iteration will also affect its performance \cite{DBLP:conf/iclr/WuDN21,9349197}. Existing practices use the whole or increasing dataset at every epoch, but non-monotonic dataset is useful for learning time series. For example, many time-sensitive applications need continuous results rather than a single-shot result. In the intensive care unit, status perception is needed at any time to rescue lives \cite{2014The}. But most detected vital signs change dynamically, leading to the evolved data distribution. This results in intertwined problems of forgetting and overfitting \cite{sunpatterns}. As shown in Figure \ref{fig:observation} (b), when learning the data at $t$, the model could forget or over-fit the data at $t-1$. In this scenario, learning samples selectively with the dynamic dataset size is potential \cite{2021A}. However, continual learning technology has not yet been targeted to time series data.

To address the above difficulties, we made some attempts and got two observations as below, which help us make a conclusion that a well-designed loss function could be a simple but effective solution. 

\begin{itemize} \setlength{\itemsep}{2pt}
    \item The loss function can discriminate samples naturally. Easy samples and hard samples behave differently in terms of their respective loss. A large loss will update the model more through backpropagation. Such discrimination makes the loss possible to represent difficulty. 

    \item The loss distribution can be changed regularly. As shown in Figure \ref{fig:observation} (c), if losses are normally distributed, they will gradually become half-normally distributed with uniform sampling; If the losses are half-normally distributed, they will gradually become normally distributed with exponential sampling. We tested cyclical sampling and get the changing loss distribution. 
\end{itemize}

Based on the observations, we design a loss to imitate the easy-to-hard process without the extra procedure such as representing the abstract time series and train the model using an adaptive cyclical dataset size to enhance its performance. 

Thus, we propose a novel CuRricUlar and CyclIcAL loss (\mname) for deep learning models to learn time series data. It follows confidence-aware losses \cite{DBLP:conf/cvpr/NovotnyALV18,DBLP:conf/cvpr/KendallGC18,DBLP:conf/nips/RevaudSHW19,DBLP:conf/nips/SaxenaTD19,DBLP:conf/nips/CastellsWR20} which can jointly learn model parameters and sample weights in a unified form. \mname schedules the learning order based on the easy-to-hard principle. It automatically and gradually upweights the contribution of the sample with large loss by modulating the loss amplitude; \mname updates the model by the dynamic training set having a cyclically changed size. We prove that the expected error between the optimal gradient and the updated gradient by using the cyclical size is smaller than that by using monotonous size. \mname is a generic representation, we give its two specific versions \mnamee and \mnameee.

\begin{figure*}[t]
\centering
\includegraphics[width=\linewidth]{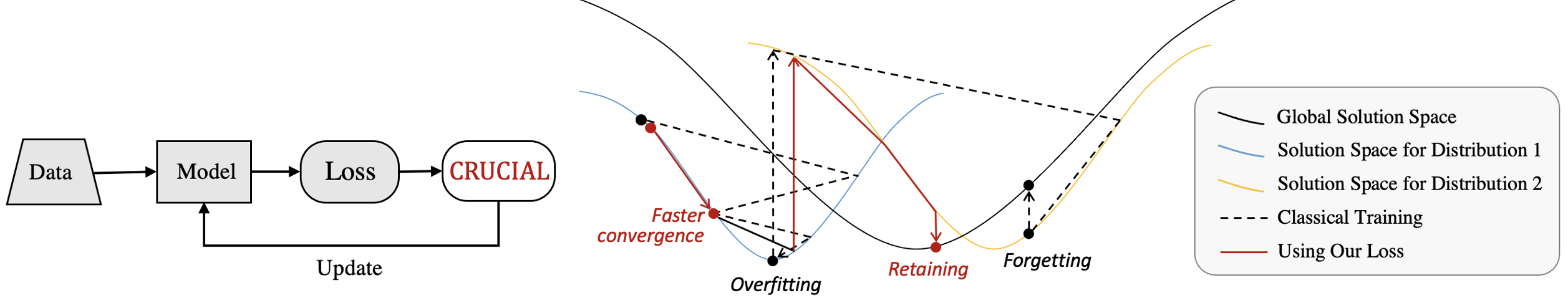}
\caption{Mechanism of \mname}
\label{fig:method}
\end{figure*}

\mname is a task-agnostic loss aiming at time series data for the first time. As shown in Figure \ref{fig:method} left, it can be plugged on top of the original loss without extra procedure. Experiments on five real-world datasets with regression and classification tasks show its benefits.

\section{Related Work}

\subsection{Schedule of Learning Order}

Making deep learning models to learn data in a meaningful order can get better results than random shuffling \cite{DBLP:journals/pami/WangCZ22}. 

A sub-discipline, curriculum learning \cite{DBLP:conf/iclr/WuDN21}, is inspired by the human learning process, naturally learning easier knowledge before more complex ones \cite{DBLP:conf/acl/ZhouYWWC20}. But many methods determine the order before learning, leading to potential inconsistencies between the fixed curriculum and the model being learned \cite{DBLP:conf/nips/CastellsWR20}. Self-paced learning \cite{DBLP:conf/nips/KumarPK10,DBLP:conf/nips/ZhouWB20} remedy this through the dynamic curriculum. But they either introduce additional parameters or rely on extra procedures like reinforcement learning. And most curricula are customized for image data, while few are for abstract time series data. The existing strategies can be summarized as:
\begin{itemize} \setlength{\itemsep}{1pt}
    \item Random learning order. It is the classical strategy, where samples are randomly sorted;
    
    \item Orderly learning order. For example, the easy-to-hard order in curriculum learning. 
\end{itemize}

\subsection{Arrangement of Dataset Size}

Resizing the dataset is important and useful in both curriculum learning and continual learning \cite{9349197}.

Classical methods train deep learning models in many epochs with several batches of dataset \cite{DBLP:conf/icml/XuT22}. Each epoch includes all data samples and the size is fixed among batches \cite{DBLP:conf/nips/0003GY20}. In curriculum learning, however, assigning the same number of samples may result in fluctuations in difficulty within a batch or not enough difference between different batches. Thus, current studies propose that the increasing size is helpful \cite{DBLP:conf/icml/Axelrod0SV20,DBLP:conf/iclr/WuDN21}. However, for the abstract time series, like distinguishing the sample's difficulty, the principle of increasing size is also hard to define \cite{sunapin}. 

Besides, it seems that the simple increment can not meet the dynamic learning task. The problem of multi-distribution usually occurs in continual learning \cite{9349197}. This sub-discipline aims to address the issue of static models incapable of adapting their behavior for new knowledge, i.e., catastrophic forgetting, by reviewing old samples \cite{2021A}. However, in its setting, the model learns a new task at every moment. The old task and new task are clear so that the multi-distribution is fixed. But the dynamic time series has data correlation over time, which easily further causes the overfitting problem. The existing strategies can be summarized as:
\begin{itemize} \setlength{\itemsep}{1pt}
    \item Full dataset. It uses all samples according to determined batch size with gradient descent; 
    \item Increasing size dataset. This strategy uses more and more samples to train the model; 
    \item Dynamic size dataset. The size changes and are not monotonous. e.g., reviewing important samples.
\end{itemize}

As analyzed above, we summarize the related mechanisms into 2 categories, where each category has 2 or 3 subcategories. They finally can be combined into 6 approaches: Random \& full; Random \& increasing; Random \& dynamic; Orderly \& full, Orderly \& increasing; Orderly \& dynamic.

However, these approaches always require important changes in the learning procedure, like multi-stage learning, extra learnable parameters, and extra procedures. Besides, the curriculum oriented to the characteristics of time series data has not been proposed. Thus, in this work, we propose a simple version, just using a loss with no additional process to achieve automatic sample perception.

\section{Methods}
{In this section, we will first define the relevant concepts to show that the effectiveness of \mname lies in two prominent characteristics of easy-to-hard order and cyclical size (Section \ref{sec:definitions}), then give the expression of \mname and its two versions \mnamee and \mnameee (Section \ref{sec:expression}), finally introduce its two mechanisms by proving its five properties (Section \ref{sec:properties}).}

\subsection{Problem Formulation}\label{sec:definitions}

\begin{definition} [Process of Learning Time Series]\label{def:task} A dataset $\mathcal{S}=\{X_{i}\}_{i=1}^{N}$ contains $N$ time series. Each time series $X=\{x_{t}\}_{t=1}^{T}$ consists of $T$ observations sorted by time. 
\begin{itemize}\setlength{\itemsep}{2pt}
    \item[-] For one-shot task, the model $f$ first learns the full-length samples $X_{1:T}$ or subsequences $X_{1:t},t<T$ with labels $Y\in \mathcal{Y}$, where $\mathcal{Y}$ can be a set of class, value or other ground truth. Then, it is optimized by minimizing loss $\frac{1}{N}\sum_{i=1}^{N}\mathcal{L}(f(X_{i}),Y_{i})$. Finally, it can give the one-shot result about the predicted label. 

    \item[-] For continuous task, the model $f$ first learns the current time series, then learns the time series with new observations, and continues over time. As time series varies among time, they forms $T$ different distributions $\mathcal{D}=\{\mathcal{D}_{t}\}_{t=1}^{T}$, $\mathcal{D}_{t}=\{X_{i,1:t}\}_{i=1}^{N}$. When the model learns $\mathcal{D}_{t}$, its performance on all observed data cannot degrade: $\frac{1}{t}\sum_{i=1}^{t}\mathcal{L}(f^{t},\mathcal{D}_{i}) \leq \frac{1}{t-1}\sum_{i=1}^{t-1} \mathcal{L}(f^{t-1},\mathcal{D}_{i})$.
\end{itemize}
\end{definition}

Multivariate time series can be described by changing $x_{t}$ to $x_{t}^{d}$, where $d$ represents the $d$-th dimension. 

\begin{definition} [Difficulty of Time Series Sample]\label{def:difficulty} The difficulty of the sample is defined by the performance of the model to this sample. The definition from easy to hard can be continuous, like a range from $0$ to $1$, where $0$ represents the easiest and $1$ represents the hardest. In this work, we give a clear division: For a time series $X_{i}$, if its loss $l_{i}$ is less than a threshold $\epsilon$, it will be an easy sample; Otherwise, it will be a hard sample. i.e., if $l_{i}<\epsilon$, then $X_{i}$ is easy; if $l_{i}\geq\epsilon$, then $X_{i}$ is hard.
\end{definition}

\begin{definition} [Confidence-aware Loss]\label{def:loss} The confidence-aware loss has the form of Formula \ref{eq:confidence-aware loss}, where $\kappa$ is the confidence coefficient of loss $l$, indicating the uncertainty of the current result. It has the natural advantage that the influence of samples on the model can be perceived or adjusted through the confidence coefficient.
\begin{equation}\label{eq:confidence-aware loss}
    L=\kappa\cdot l
\end{equation}
\end{definition}

\begin{definition} [Curricular and Cyclical Loss \mname]\label{def:ccl} Based on the form of confidence-aware loss in Definition \ref{def:loss}, \mname has two additional points: arranging an easy-to-hard learning order of samples and scheduling a cyclical size of the dataset. These two properties should be realized from coefficient constraint $\kappa$ by modulating the loss amplitude.
\end{definition}

\begin{table}[!t] 
\centering
\caption{{Notations}}\label{tb:notations}
\begin{tabular}{ll}
\toprule[0.8pt]
Notation  &Description \\
\midrule[0.8pt]
$X_{i},x_{i,t}, Y_{i}$ & A time series, its value at time $t$, and its label\\
$\mathcal{S},\mathcal{D}$ & Dataset and its distribution \\
$f,w$ & Deep learning model and its parameters\\
\midrule[0.4pt]
$\mathcal{L}$ & Overall loss of a task\\
$l_{i}$ & Original loss when learning sample $X_{i}$\\
$L_{i}$ & New loss of $X_{i}$ after using \mname\\
$\kappa$ & Confidence coefficient to modulate $l$\\
$\lambda$ &  Regularization coefficient to constrain $\kappa$\\
$\epsilon$ &  Threshold to distinguish difficulty of $l$\\
$\varepsilon$ & Indication of current training status\\
$\varphi_{C}$ &Constraint function about $\kappa$\\
$\varphi_{D}$ &Differentiation function about $\epsilon$\\
$\varphi_{S}$ &State function about $\varepsilon$\\
$\mathcal{N}(\mu_{l},\sigma^{2}_{l})$ & Distribution of loss $l$ (mean and variance)\\
\bottomrule[0.8pt]
\end{tabular}
\end{table}

\subsection{Expression of \mname}\label{sec:expression}

\subsubsection{{General expression}}

Based on Definition \ref{def:ccl}, we expect to dynamically control the loss confidence of samples, making the model learn different samples at different stages. i.e., the loss should provide an easy-to-hard order and a cyclical size. 

To this end, we first establish the general expression of \mname in Equation \ref{eq:general formula}. $\kappa$ and $\epsilon$ are responsible for making the loss curricular and cyclical respectively. 
\begin{equation} \label{eq:general formula}
L=\kappa \cdot \varphi_{D}(l,\epsilon)+\varphi_{C}(\kappa), \quad \epsilon=\varphi_{S}(\varepsilon)
\end{equation}

About curriculum, after being operated by the constraint function $\varphi_{C}$ with the parameter $\lambda$, the confidence coefficient $\kappa$ can modulate the loss amplitudes for different samples, which are distinguished by the differentiation function $\varphi_{D}$ with the threshold $\epsilon$.

About cycle, the threshold $\epsilon$ is changed periodically according to the current training state $\varepsilon$ of the model and represented by the state function $\varphi_{S}$ about $\varepsilon$, and finally controls the size of the current dataset. 

Thus, to get the curricular and cyclical outcome, the key is to design three functions in \mname:
\begin{itemize}\setlength{\itemsep}{2pt}
    \item Constraint function $\varphi_{C}$;
    \item Differentiation function $\varphi_{D}$;
    \item State function $\varphi_{S}$.
\end{itemize}

In this work, through designing these three equations, we propose a loss version having the manual cycle \mnameee (Section \ref{sec:SIN}) and a loss version having the adaptive cycle \mnamee (Section \ref{sec:ADP}).

\subsubsection{Manual cycle \mnamee} \label{sec:SIN}
{Intuitively, we can manually control the cycle of \mname loss. Thus, three key functions are expressed as:}
\begin{equation}\label{eq:sin functions}
\begin{aligned}
\varphi_{C}(\kappa)=\lambda&\cdot(\log\kappa)^{2},\ \lambda=-\log{\mathcal{F}}\\
&\varphi_{D}(l,\epsilon)=l-\epsilon\\
\varphi_{S}(\varepsilon)=&2\varepsilon- \mu_{l},\ \varepsilon=\frac{1}{2}\mathcal{F}\cdot \mu_{l}
\end{aligned}
\end{equation}

{To achieve the cyclical size, we define $\varepsilon$ and $\lambda$ by the periodic function $\mathcal{F}$, and make $\epsilon$ in the same rhythm with $\varepsilon$ based on the base threshold $\mu_{l}$, where $\mu_{l}$ is the average of all losses. $\mathcal{F}$ is optional and can be }
\begin{equation}\label{eq:f}
\mathcal{F}=\sin^{2}{(\omega t+\mathcal{T})}, t\in \mathbb{N}
\end{equation}

Thus, the specific representation of \mnamee is: 
\begin{equation}
\begin{aligned}
    L_{i,t}&=\text{\mnamee}(l_{i,t})\\
    &=\left\{ \begin{aligned} 
    \kappa_{i,t}\cdot(l_{i,t}-\epsilon_{t})&+\lambda_{t}\cdot(\log\kappa_{i,t})^{2} \ \ l_{i,t}\geq \varepsilon_{t}\\
    &0 \quad\quad\quad\quad\quad\quad\ \quad l_{i,t}<\varepsilon_{t}
    \end{aligned} \right.
\end{aligned}
\end{equation}

\subsubsection{{Adaptive cycle \mnameee}} \label{sec:ADP}

\mnamee still requires defining periodic hyper-parameters, making it difficult to match datasets and models empirically. Thus, we propose an adaptive version \mnamee. Three key functions loss are expressed as:
\begin{equation} \label{eq:adp functions}
\begin{aligned}
\varphi_{C}(\kappa)=\lambda\cdot&(\log\kappa)^{2}\\
\varphi_{D}(l,\epsilon)=& l-\epsilon\\
\varphi_{S}(\varepsilon)=\varepsilon\cdot\mu_{l},\  \varepsilon=\textsc{sk}&=\frac{\sum_{i}^{N}(l_{i}-\mu_{l})^{3}}{N\sigma_{l}^{3}}
\end{aligned}
\end{equation}

\begin{itemize}\setlength{\itemsep}{3pt}
    \item Constraint function $\varphi_{C}$. To define the confidence coefficient $\kappa$, we utilize a regularization term $\lambda(\log\kappa)^{2}$. In this way, we can use the method of closed-form formula \cite{DBLP:conf/nips/CastellsWR20} to obtain the final value $\kappa^{*}=\arg\min_{\kappa}L(l,\kappa)$, where we directly use their converged value at the limit, which only depends on the input loss $l$, without requiring additional parameters or inducing a convergence delay. Finally, $\kappa_{i,t}=e^{-\frac{1}{2}W\max(-\frac{2}{e},\frac{l_{i,t}-\epsilon_{t}}{\lambda})}$ in Equation \ref{eq:kappa}. 

    \item Differentiation function $\varphi_{D}$. To distinguish samples as described in Definition \ref{def:difficulty}, we directly adopt the difference between the sample loss and the threshold. For example, the current sample $X_{i}$ is an easy sample when the difference between its loss $l_{i}$ and the average of all losses $\mu_{l}$ is negative. 
    
    \item State function $\varphi_{S}$. To realize the adaptive cycle, we determine the samples selected in the current round $t$ according to the current model state $\varepsilon$. $\varepsilon$ is reflected in the loss distribution of all samples $\mathcal{N}(\mu_{t},\sigma_{l_t}^{2})$. Based on the observation in Figure \ref{fig:observation} and proof in Section \ref{sec:properties}, this distribution changes between the normal distribution and the half-normal distribution. Thus, we use the skewness coefficient \textsc{sk} in Equation \ref{eq:epsilon} to quantify.
\end{itemize}

{Thus, the specific representation of \mnameee is: }
\begin{equation} \label{eq:ccl}
\begin{aligned}
    L_{i,t}&=\text{\mnameee}(l_{i,t})\\
    &=\kappa_{i,t}\cdot(l_{i,t}-\epsilon_{t})+ \lambda\cdot(\log\kappa_{i,t})^{2}
\end{aligned}
\end{equation}

\begin{equation} \label{eq:kappa}
    \text{where }\ \kappa_{i,t}=e^{-\frac{1}{2}W\max(-\frac{2}{e},\frac{l_{i,t}-\epsilon_{t}}{\lambda})} \quad \
\end{equation}

\begin{equation} \label{eq:epsilon}
    \text{and }\ \epsilon_{t}=\frac{\sum_{i}^{N}(l_{i,t-1}-\mu_{l_{t-1}})^{3}\cdot \mu_{l_{t-1}}}{N\sigma_{l_{t-1}}^{3}}
\end{equation}

{Note that the expression of $\kappa$ applies to both \mnamee and \mnameee, where $W(*)$ is the Lambert W function for the solution to $y=xe^{x}$. Refer to Appendix \ref{app:expression} for the deduction.}

\begin{figure*}[t]
\centering
\includegraphics[width=\linewidth]{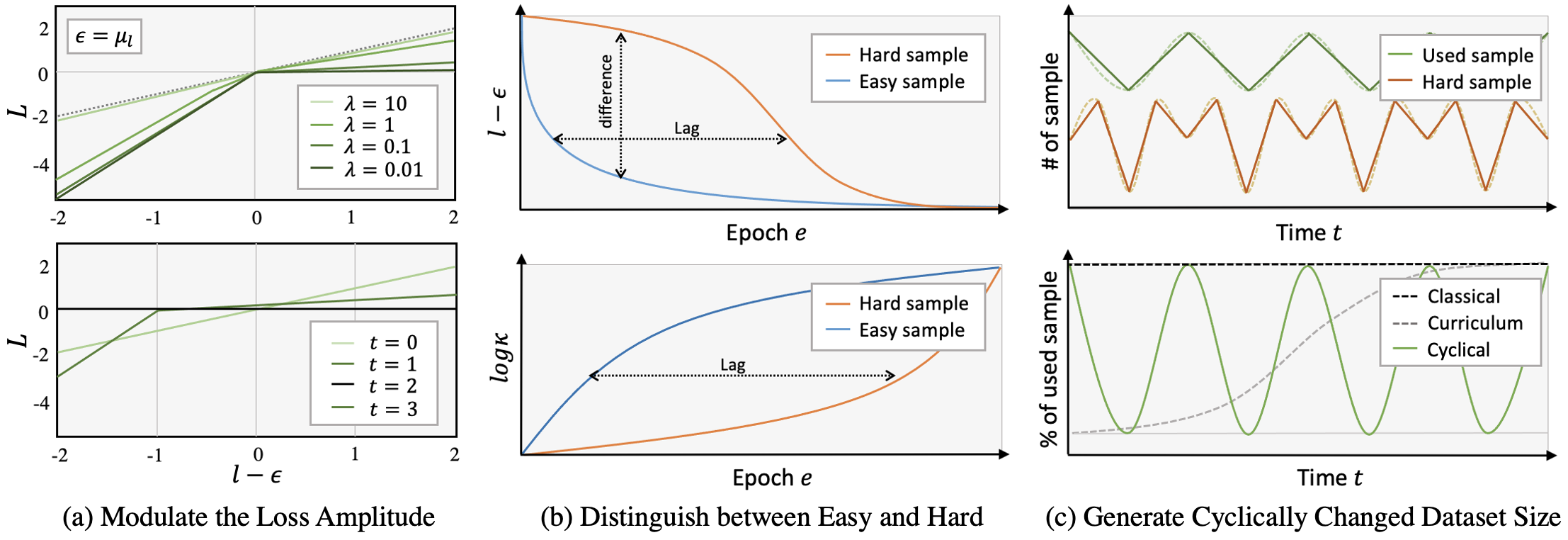}
\caption{{Properties of \mname}}
\label{fig:property}
\end{figure*}

\subsection{Property of \mname} \label{sec:properties}

\subsubsection{Basic properties}

\mname loss function satisfy three basic properties for generality \cite{DBLP:conf/nips/CastellsWR20}. 

\begin{property}[Translation-invariance]\label{prop:1}
$\forall C, \exists C'|L(l+C,\kappa)=L(l,\kappa)+C'$. $C$, $C'$ are constant.
\end{property}
\begin{property}[Homogeneity]\label{prop:2}
$\exists \lambda, \lambda'|\forall C>0, L_{\lambda'}(C\cdot l,\kappa)=C\cdot L_{\lambda}(l,\kappa)$. $C$ is a constant.
\end{property}
\begin{property}[Generalization]\label{prop:3}
$\exists \kappa | L(l,\kappa)=l+C$. $C$ is a constant.
\end{property}

With Property \ref{prop:1}, if we add a constant to the input, it should have no effect on the loss’s gradient; With Property \ref{prop:2}, we can accordingly rescale $\lambda$ to handle input losses of any amplitude; With Property \ref{prop:3}, it can amount to the input loss for a particular confidence. Refer to Appendix \ref{app:proof} for proofs.

\subsubsection{Characteristic properties}

After satisfying the basic properties, \mname satisfies two additional characteristics defined in Definition \ref{def:ccl}. The curricular learning order is achieved by Property \ref{prop:4} and the cyclical dataset size is achieved by Property \ref{prop:5}. They are the main points of our method.

\begin{property}[Differentiated Scaling]\label{prop:4}
$\forall l_{i},l_{j},l_{i}<\epsilon,l_{j}>\epsilon|\frac{L_{i}}{l_{i}-\epsilon}>\frac{L_{j}}{l_{j}-\epsilon}$
\end{property}

\begin{figure*}[t]
\centerline{
\includegraphics[width=1.03\linewidth]{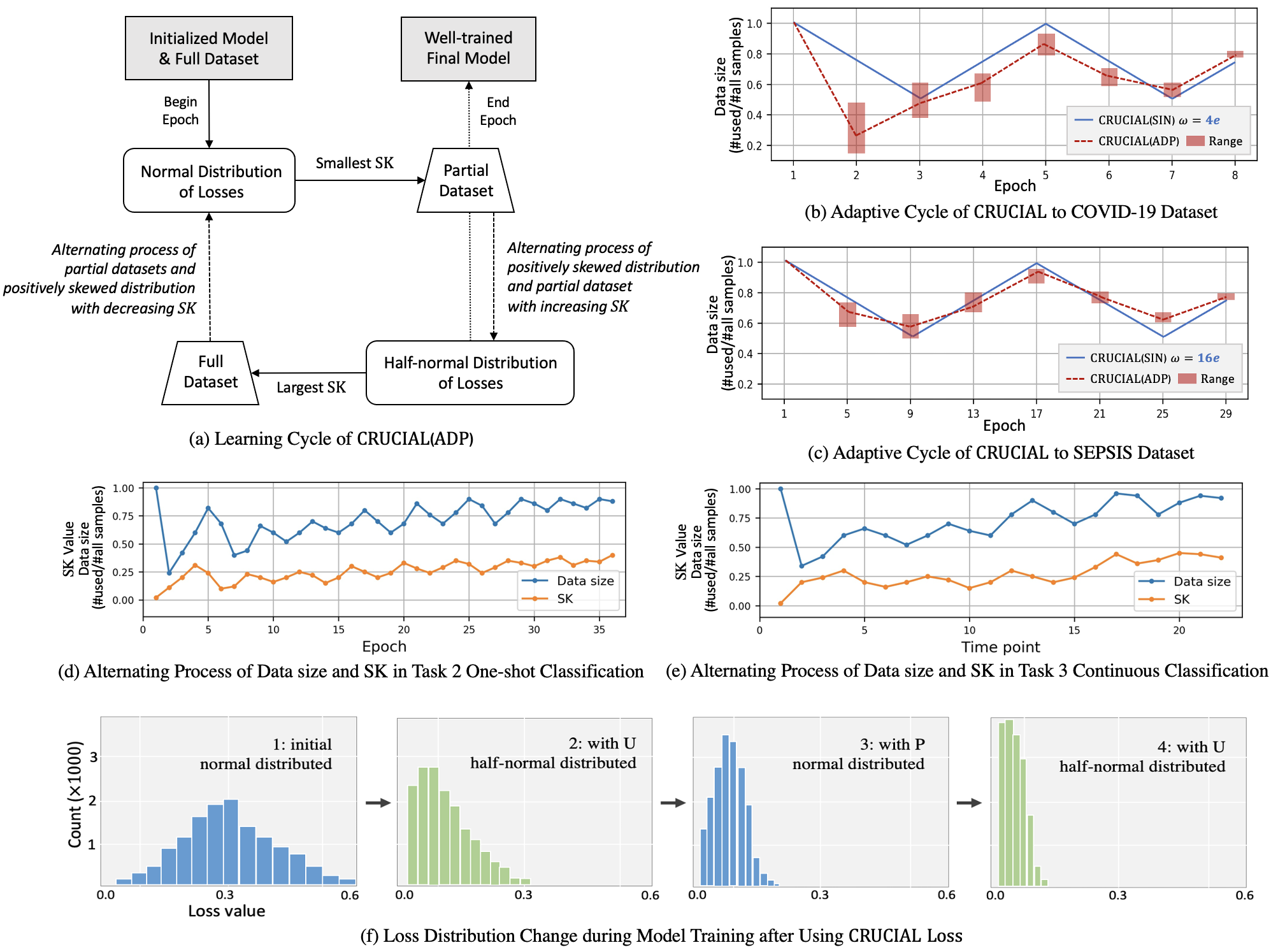}}
\caption{{Cycle of \mnameee}}\label{fig:cycle}
\end{figure*}

{In \mname, $\epsilon$ is a threshold that ideally separates easy samples from hard samples based on their respective loss. For the selected samples, \mname will change their losses to varying degrees: taking $\epsilon$ as the boundary, the losses of hard samples ($l>\epsilon$) will be scaled down, and the that of easy samples ($l<\epsilon$) will be expanded as shown in Figure \ref{fig:property} (a), i.e., it amplifies the contribution of easy samples than hard ones. The model is affected by samples with higher gradients. Which means that the simple samples participate more in the early training than the hard samples.}

{With the increase of epochs and the convergence of the model, the hard samples become easier, thus the model is affected more by them. It is consistent with the easy-to-hard learning order of curriculum learning. As shown in Figure \ref{fig:property} (b), we plot the typical losses incurred by an easy and a hard sample at the top. We show their respective confidence when learned via back-propagation using our loss at the bottom. This process induces a significant delay between the moment a hard sample becomes easy. In the context of confidence-aware loss, the confidence $\log\kappa$ of the hard sample lags behind that of the easy sample. }

\begin{property}[Cyclical Sampling with Less Total Error]\label{prop:5}
$\exists\varepsilon, \forall C| \varepsilon_{t} \approx\varepsilon_{t+p}, \sum_{i=1}^{N}L_{\varepsilon}(l_{i})<\sum_{i=1}^{N}L_{C}(l_{i})$, $p$ is the period and $C$ is the constant threshold.
\end{property}

This property shows that the cyclical size of dataset can make the model achieve better performance. We will prove it according to the following Theorem \ref{theo:1}, \ref{theo:2}, and \ref{theo:3}.

The non-monotonic dataset size is needed due to the intertwined problem of forgetting and overfitting when continuously learning time series over time, where the model learns subsets $\mathcal{S}_{t}=\{X_{i,1:t}\}_{i=1}^{N}$ one by one from $\mathcal{S}$. Due to uncertain similarity and difference between adjacent $X_{1:t}$ and $X_{1:t-1}$, we arrange the size from the non-full/partial dataset, specifically. We prove that in the learning process, the expected error between the updated gradient and optimal gradient by using the alternate data size is smaller than that by using monotonous ones. The process is Theorem \ref{theo:1} $+$ Theorem \ref{theo:2} $\rightarrow$ Theorem \ref{theo:3}. 

A deep learning model $f$ with parameters $w$ try to minimize the loss $l_{i}$ of each input sample $X_{i}$ throughout training $\mathcal{L}=\frac{1}{N}\sum^{N}_{i=1}l_{i}$. The optimization process can be performed with the gradient descent algorithm $w_{t+1}=w_{t}-\eta\nabla_{w}\mathcal{L}(w_{t})$. But in practice, the optimization is performed based on data batch, like (momentum) stochastic gradient descent algorithm \cite{DBLP:conf/nips/0003GY20}, $w_{t+1}=w_{t}-\gamma\nabla_{w}l_{i}(w_{t})$. 

Based on them, the expected value of mean squared error between updating the model using partial samples and all samples is $ E[\Delta(w_{t})]=E[(\nabla_{w}l_{i}(w_{t})-\nabla_{w}L(w_{t}))^{2}]=E[(\nabla_{w}(l_{i}(w_{t})-\mathcal{L}(w_{t})))^{2}]$. Thus, at the training time $t$, the main factor leading to the difference between updating $w_{t}$ to $w_{t+1}$ and to optimal $\overline{w}_{t+1}$ is reflected in the mean squared error between the loss $l_{i}$ and the global loss $\mathcal{L}$.
\begin{equation}
\begin{aligned}
    E=E[(l_{i}(w_{t})-\mathcal{L}(w_{t}))^{2}]
\end{aligned}
\end{equation}

During the process of updating model parameters, we focus on two conditions defined below. \textsc{Condition U}: If training samples are randomly selected from a uniform distribution, the $i$-th sample will be selected with the mean probability $\frac{1}{N}$; \textsc{Condition P}: If training samples are selected differently, the model training process will be the exponential distributed stochastic gradient descent. For example, in \mnameee, samples whose $l_{i}<\epsilon$ are more likely to be selected as their loss amplitudes are maintained but that of samples whose $l_{i}>\epsilon$ are modulated smaller. As the losses are uniformly distributed, the probability is exponentially distributed with $p(l_{i})=\Lambda e^{-\Lambda l_{i}}$. $\Lambda$ is the rate parameter. The $i$-th sample will be selected with the exponential probability. 
\begin{equation}
\begin{aligned}
\textsc{Condition U: } &\text{Select samples randomly} \Rightarrow \\
    &X_{i} \text{ is selected with probability } \frac{1}{N}\\
\textsc{Condition P: } &\text{Select samples } \text{ based on loss } l_{i} \Rightarrow \\
    X_{i} &\text{ is selected with probability }\frac{\Lambda e^{-\Lambda l_{i}}}{\sum_{i=1}^{N}\Lambda e^{-\Lambda l_{i}}}
\end{aligned}\nonumber
\end{equation}

Based on above two conditions, we propose three theorems. \mname satisfies Theorem \ref{theo:1} naturally, and satisfies the condition $\sigma<\frac{\pi}{\Lambda}$ in Theorem \ref{theo:2} when we set $\lambda<0.01$ in the constraint function $\varphi_{C}$. Theorem \ref{theo:1} and \ref{theo:2} are the foundation of Theorem \ref{theo:3}. Refer to Appendix \ref{app:proof} for their detailed proofs.

\begin{theorem} \label{theo:1}
At learning time $t$, if the loss values are normal distributed $\mathcal{N}(\mu,\sigma^{2})$, the expected error of U will be lower than that of P:
\begin{equation}
    l_{i}(w_{t})\sim \mathcal{N}(\mu,\sigma^{2})\Longrightarrow E_{U}<E_{P}
\end{equation}
\end{theorem}

\begin{theorem} \label{theo:2}
At learning time $t$, if the loss values are half-normal distributed $\mathcal{N}(\mu+\sigma\sqrt{\frac{2}{\pi}},\sigma^{2}(1-\frac{2}{\pi}))$, the expected error of P will be lower than that of U under the condition of $\sigma<\frac{\pi}{\Lambda}$:
\begin{equation}
    l_{i}(w_{t})\sim \mathcal{N}(\mu+\sigma\sqrt{\frac{2}{\pi}},\sigma^{2}(1-\frac{2}{\pi}))\land\sigma<\frac{\pi}{\Lambda}\Longrightarrow E_{P}<E_{U}
\end{equation}
\end{theorem}

Based on the fact in Figure \ref{fig:observation} (c), demonstrating that if losses come from the normal distribution, they will gradually become half-normal distributed with uniform sampling, and if the losses come from the half-normal distribution, they will gradually become normally distributed with exponential sampling \cite{DBLP:journals/corr/abs-2202-05531}, Theorem \ref{theo:1} and \ref{theo:2} can appear alternately, implementing a cyclical process. 

Thus, we come to the conclusion in Theorem \ref{theo:3}: If we train the model cyclically using a uniform sampled dataset and an exponential sampled dataset, the loss will be lower than that using a single dataset. 

\begin{theorem}\label{theo:3}
The expected error of classical loss becomes lower with the constraint of \mname.
\begin{equation}
    E_{\text{\mname}}<E
\end{equation}
\end{theorem}

In \mnameee, the selection of samples is based on the modulated loss $L$ of each sample. Where the threshold $\epsilon$ controls which sample losses will be modulated and is determined by the skewness coefficient \textsc{sk}. \textsc{sk} is the the driving factor of the whole cycle in Figure \ref{fig:cycle} (a): 

{Input full dataset into initialized model $\rightarrow$ $\bullet$ The loss set is like normal distribution $\rightarrow$ Smaller \textsc{sk} $\rightarrow$ More loss amplitudes are decreased so as not to affect the model $\rightarrow$ Condition P $\rightarrow$ Smaller/partial dataset $\rightarrow$ The loss set is like half-normal distribution $\rightarrow$ Larger \textsc{sk} $\rightarrow$ More loss amplitudes are maintained to affect the model $\rightarrow$ Condition U $\rightarrow$ Full dataset $\rightarrow$ $\bullet$ The loss set is like normal distribution ($\bullet$ indicates the beginning and the end of the cycle).}

{This cycle process presented after using \mnameee is spontaneous, self-consistent, and adaptive. It does not need to be driven by the additional periodic function.}

\begin{table*}[!t]
\caption{Task \ref{task:1} Regression Results (MSE$\downarrow$) of Three Base Models with Different Strategies and Losses\\
1 is random order, 2 is easy-to-hard order; I is random data size, II is curriculum data size.}\label{tb:regression results}
\resizebox{\textwidth}{!}{
\begin{tabular}{l|l|ccc|ccccc}
\toprule
 &Model  &1\&II &2\&I &2\&II &Original &WeightLoss &SuperLoss  &\mnamee &{\textbf{\mnameee}}\\
\midrule
\multirow{3}*{SILSO} &LSTM &2.61±0.59   &2.57±0.73 &2.84±0.69  &2.95±0.74 &2.73±0.81 &2.56±0.69  &2.39±0.78 &\textbf{2.32±0.79}\\
&CNN &2.93±0.55   &2.98±0.71 &3.23±0.44  &3.57±0.64 &2.88±0.56 &2.79±0.41  &2.60±0.43 &\textbf{2.55±0.61}\\
&Transformer &3.09±0.95   &2.97±1.01 &3.35±0.78  &3.60±0.82 &2.97±0.96 &2.94±0.75  &\textbf{2.85±0.84} &2.89±0.91\\
\midrule
\multirow{3}*{USHCN}  &LSTM &0.69±0.20 &0.66±0.12 &0.73±0.15 &0.75±0.32 &0.74±0.33 &0.68±0.25  &0.64±0.19 &\textbf{0.63±0.17}\\
 &CNN &0.75±0.24 &0.69±0.14 &0.77±0.21 &0.75±0.33 &0.74±0.31 &0.70±0.27  &0.66±0.21 &\textbf{0.62±0.20}\\
 &Transformer &0.79±0.55 &0.79±0.36 &0.83±0.65 &0.79±0.69 &0.77±0.52 &0.74±0.40  &0.70±0.39 &\textbf{0.69±0.42} \\
\bottomrule
\end{tabular}
}
\end{table*}

\begin{table*}[!t]
\caption{Task \ref{task:2} Classification Accuracy (AUC-ROC$\uparrow$) of Different Strategies\\
All strategies and losses are based on LSTM model; Task for COVID-19* is early 3 days classification; Task for SEPSIS* is early 6 hours classification; Others are classical classification tasks with full-length data.}\label{tb:classification accuracy 2}
\resizebox{\textwidth}{!}{
\begin{tabular}{l|ccc|ccccc}
\toprule[0.8pt]
   &1\&II &2\&I &2\&II &Original &WeightLoss &SuperLoss  &\mnamee &{\textbf{\mnameee}}   \\
\midrule[0.8pt]
UCR-EQ   &0.935±0.004 &0.937±0.004 &0.936±0.007 &0.934±0.005 &0.938±0.004 &0.939±0.004  &\textbf{0.942±0.003} &0.941±0.004\\
USHCN  &0.930±0.010  &0.945±0.010 &0.932±0.010 &0.925±0.013 &0.930±0.014 &0.934±0.012 &0.937±0.013 &\textbf{0.939±0.017}\\
COVID-19  &0.991±0.005  &0.992±0.004 &0.987±0.005  &0.988±0.003 &0.990±0.004 &0.993±0.002 &\textbf{0.996±0.001}  &0.994±0.003 \\
COVID-19* &0.963±0.005  &0.960±0.007 &0.960±0.006  &0.950±0.008 &0.962±0.007 &0.965±0.009 &0.968±0.010 &\textbf{0.969±0.010} \\
SEPSIS   &0.899±0.011  &0.901±0.010 &0.885±0.009  &0.886±0.010 &0.900±0.011  &0.915±0.010 & 0.928±0.011 & \textbf{0.929±0.012}\\
SEPSIS*   &0.859±0.012  &0.867±0.012 &0.865±0.012  &0.855±0.017 &0.863±0.012  &0.868±0.017 & 0.871±0.012 & \textbf{0.873±0.014}\\
\bottomrule[0.8pt]
\end{tabular}}
\end{table*}

\section{Experiments}
 The experiments show that our method can be used for different tasks and models (Section \ref{sec:experiment setting}). It outperforms baselines (Section \ref{sec:results}), and has some unique properties and characteristics (Section \ref{sec:analysis}).

\subsection{Tasks, Datasets, and Baselines} \label{sec:experiment setting}

\begin{task}[Regression]\label{task:1}
Regression tasks require to forecast the future value after learning the existing full-length or partial-length time series. We use SILSO dataset \cite{SILSO} to one-step-ahead forecast the monthly sunspot and use USHCN dataset \cite{USHCN} to predict New York temperature. We straightforwardly plug \mname on top of the Mean-Square-Error (MSE) loss. The loss at $e$-th epoch becomes:
\begin{equation}
   \mathcal{L}_{task1}=\frac{1}{N}\sum_{i=1}^{N}(\kappa_{i,e}(l^{mse}_{i,e}-\epsilon_{e})+\lambda(\log\kappa_{i,e})^{2}) 
\end{equation}
\end{task}

\begin{task}[Single-shot Classification]\label{task:2}
Classical classification tasks require to classify time series after learning the full-length data. We use the UCR-EQ dataset \cite{UCRArchive} to predict the earthquakes; Early classification tasks classify time series after learning the subsequences in the early stage. We use SEPSIS dataset \cite{DBLP:conf/cinc/ReynaJSJSWSNC19} to 6-hour early diagnosis. We plug \mname on top of the Cross-Entropy (CE) loss $l^{ce}_{i}$. Then, the task loss at $e$-th epoch becomes:
\begin{equation}
   \mathcal{L}_{task2}=\frac{1}{N}\sum_{i=1}^{N}(\kappa_{i,e}(l^{ce}_{i,e}-\epsilon_{e})+\lambda(\log\kappa_{i,e})^{2}) 
\end{equation}
\end{task}

\begin{task}[Continuous Classification]\label{task:3}
Continuous classification tasks require to classify time series at every time although the class of a time series is usually labeled at the final time. We use the COVID-19 dataset \cite{COVID-19} to predict mortality at every time point in time series \cite{DBLP:journals/BMC/sun} and use the SEPSIS dataset \cite{DBLP:conf/cinc/ReynaJSJSWSNC19} to diagnosis sepsis at every time point. We also plug \mname on top of the Cross-Entropy (CE) loss. But not at epoch $e$, the task loss at $t$-th time point in the time series becomes:
\begin{equation}
     \mathcal{L}_{task3}=\frac{1}{N}\sum_{i=1}^{N}(\kappa_{i,t}(l^{ce}_{i}-\epsilon_{t})+\lambda(\log\kappa_{i})^{2}) 
\end{equation}
\end{task}

The regression accuracy in Task \ref{task:1} is evaluated by Mean Square Error (MSE); The classification accuracy in Task \ref{task:2}, \ref{task:3} is evaluated by Area Under Curve of Receiver Operating Characteristic (AUC-ROC). For continuous tasks in Task \ref{task:3}, performances of solving forgetting and overfitting problems are evaluated by backward transfer \textsc{bwt} and forward transfer \textsc{fwt}, which evaluate the influence that learning current data has on the old/future data. $R\in \mathbb{R}^{|\mathcal{D}|\times |\mathcal{D}|}$ is an accuracy matrix, $R_{i,j}$ is the accuracy on $\mathcal{D}_{j}$ after learning $\mathcal{D}_{i}$. $\overline{b}$ is the accuracy with random initialization.
\begin{equation} 
    \textsc{bwt}=\sum_{i=1}^{|\mathcal{D}|-1} \frac{R_{|\mathcal{D}|,i}-R_{i,i}}{|\mathcal{D}|-1}
\end{equation}
\begin{equation} 
    \textsc{fwt}=\frac{\sum_{i=2}^{|\mathcal{D}|}R_{i-1,i}-\overline{b}_{i,i}}{|\mathcal{D}|-1}
\end{equation}

\mname is the model-agnostic loss function. In addition to testing its ability to decouple from tasks and datasets, we also used it for different models. We adopt 3 deep neural networks as the base model. On this basis, we implement different strategies and losses for comparison.

\begin{itemize}\setlength{\itemsep}{2pt}
    \item Base models
        \begin{itemize}
            \item LSTM \cite{DBLP:journals/tai/GuptaGBD20,Baytas:2017:PSV:3097983.3097997}, based on recurrent neural networks, models time series recursively and protects long short-term dependencies by some special gates;
            \item CNN \cite{DBLP:journals/nn/0058S21,DBLP:journals/datamine/FawazFWIM19} extracts hierarchical patterns in time series using stacked trainable small filters or kernels among time ranges; 
            \item Transformer \cite{DBLP:conf/aaai/ZhouZPZLXZ21,DBLP:conf/kdd/ZerveasJPBE21}, having encoder-decoder structure, uses multi-head attention to model long-range dependencies and interactions in sequential data.
        \end{itemize}
    \item Strategies for planning learning order
        \begin{itemize}
            \item Random \cite{DBLP:conf/nips/0003GY20} order is widely used, usually applying the stochastic gradient descent algorithm \cite{DBLP:conf/nips/0003GY20};
            \item Easy-to-hard \cite{DBLP:conf/iclr/WuDN21} order lets the model learn easy samples then hard samples. 
        \end{itemize}
    \item Strategies for planning dataset size
        \begin{itemize}
            \item Full size \cite{DBLP:journals/ijon/RizkA19} is classical and widely used, where all samples are used to train the model at each epoch; 
            \item Increasing size \cite{DBLP:conf/iclr/WuDN21} has the increased number of samples, where more and more samples are used to train the model. It has been shown as the common point of curriculum learning.
        \end{itemize}
    \item Loss functions
        \begin{itemize}
            \item WeightLoss \cite{2021A} learns task-instance parameters to tag important samples and replays them;
            \item SuperLoss \cite{DBLP:conf/nips/CastellsWR20} automatically downweights the contribution of samples with a large loss.
        \end{itemize}
\end{itemize}

\begin{table*}[!t]
\caption{Task \ref{task:3} Continuous Classification Accuracy (AUC-ROC$\uparrow$) of Three Base Models with Different Losses \\
k\% means that the current classification is based on the k\%-length time series; + means plugging the additional loss to the base model}\label{tb:classification accuracy 1}
\resizebox{\textwidth}{!}{
\begin{tabular}{c|l|ccccccc}
\toprule[0.8pt]
 &Method    &40\%  &50\%   &60\% &70\%  &80\%    &90\%   &100\%  \\
\midrule[0.8pt]
\multirow{15}*{\shortstack[c]{C\\O\\V\\I\\D\\-\\1\\9}}
& LSTM    &0.895±0.014  &0.916±0.013 &0.925±0.013     &0.941±0.012 &0.944±0.011   &0.948±0.007 &0.950±0.008       \\
&  +WeightLoss     &0.896±0.013  &0.916±0.011 &0.926±0.012     &0.945±0.013 &0.944±0.014   &0.952±0.005 &0.961±0.008       \\
&   +SuperLoss    &0.899±0.015  &0.915±0.014 &0.923±0.012     &\textbf{0.946±0.014} &0.955±0.017   &0.960±0.005 &0.963±0.005       \\
&+\mnamee &\textbf{0.901±0.015}  &\textbf{0.919±0.015}   &0.927±0.012 &0.945±0.011   &0.960±0.011   &0.965±0.010  &0.967±0.010 \\
&{\textbf{+\mnameee}} &0.900±0.013  &0.916±0.014  &\textbf{0.930±0.014}   &0.945±0.012   &\textbf{0.962±0.012}   &\textbf{0.967±0.010}  &\textbf{0.969±0.012}\\
\cline{2-9}

\rule{0pt}{10pt}
& CNN      &0.879±0.016  &0.915±0.014  &0.926±0.014       &0.936±0.010 &0.941±0.006 &0.947±0.006   &0.952±0.006       \\
&  +WeightLoss    &0.887±0.013  &0.915±0.013 &0.925±0.015     &0.935±0.009 &0.940±0.007  &\textbf{0.950±0.005}   &0.954±0.006       \\
&  +SuperLoss       &0.885±0.012  &0.912±0.011    &0.924±0.018   &0.932±0.011   &0.939±0.005  &0.943±0.004  &0.955±0.005      \\
& +\mnameee      &\textbf{0.895±0.016}   &0.915±0.014 &0.927±0.011       &0.941±0.010 &\textbf{0.943±0.007}   &0.948±0.004 &0.957±0.008       \\
&\textbf{+\mnameee} &0.894±0.018   &\textbf{0.917±0.014} &\textbf{0.930±0.014}       &\textbf{0.941±0.011} &0.942±0.006   &\textbf{0.950±0.005} &\textbf{0.958±0.007} \\
\cline{2-9}

\rule{0pt}{10pt}
& Transformer     &0.898±0.018 &0.913±0.013   &0.923±0.010  &0.937±0.007    &0.946±0.006  &0.950±0.004   &0.952±0.005   \\
&  +WeightLoss    &0.901±0.014 &0.913±0.012  &\textbf{0.924±0.015}   &0.940±0.012 &0.948±0.007  &0.958±0.008  &0.953±0.007   \\
&  +SuperLoss      &0.898±0.018   &0.916±0.017 &0.923±0.014     &0.936±0.012 &0.940±0.013   &0.955±0.009  &0.956±0.008      \\
& +\textbf{+\mnameee}   &\textbf{0.903±0.021}   &\textbf{0.917±0.015}   &\textbf{0.924±0.017} &0.942±0.011  &\textbf{0.944±0.010}  &\textbf{0.963±0.007}  &\textbf{0.964±0.005}      \\
&\mnameee &0.900±0.020   &\textbf{0.917±0.017}   &\textbf{0.924±0.017} &\textbf{0.944±0.013}  &\textbf{0.944±0.012}  &0.960±0.005  &0.963±0.005 \\

\midrule[0.4pt]
\multirow{15}*{\shortstack[c]{S\\E\\P\\S\\I\\S}}
& LSTM      &0.761±0.020   &0.822±0.023     &0.833±0.028      &0.836±0.015 &0.840±0.016  &0.845±0.016    &0.855±0.017    \\
&  +WeightLoss    &0.767±0.020   &0.827±0.022     &0.834±0.025      &0.839±0.011 &0.848±0.012  &0.849±0.016    &0.865±0.013    \\
&  +SuperLoss   &\textbf{0.768±0.021}   &0.825±0.023     &0.830±0.022      &0.837±0.015 &0.846±0.014  &0.848±0.017     &0.868±0.012    \\
&+\mnamee    &\textbf{0.768±0.026}   &0.831±0.023       &\textbf{0.840±0.024}    &0.840±0.018  &0.851±0.012  &0.853±0.012    &0.872±0.012  \\
&\textbf{+\mnameee} &0.767±0.024   &\textbf{0.833±0.024}       &0.838±0.025    &\textbf{0.840±0.019}  &\textbf{0.853±0.015}  &\textbf{0.855±0.013}    & \textbf{0.873±0.013} \\
\cline{2-9}

\rule{0pt}{10pt}
& CNN   &0.755±0.027    &0.815±0.025   &0.832±0.027   &0.830±0.028      &0.844±0.012 &0.847±0.010      &0.848±0.016    \\
&  +WeightLoss    &0.759±0.028     &0.824±0.029   &0.835±0.027   &0.838±0.026   &0.846±0.015    &0.850±0.017    &0.857±0.018    \\
&  +SuperLoss     &0.761±0.021    &0.820±0.020     &0.837±0.022  &0.832±0.021  &0.835±0.018    &0.845±0.015   &0.850±0.011    \\
&  +\mnamee   &0.765±0.025   &0.826±0.027    &0.839±0.027  &\textbf{0.839±0.025}      &\textbf{0.849±0.017} &0.852±0.011      &\textbf{0.865±0.016}    \\
&\textbf{+\mnameee} &\textbf{0.766±0.026}    &\textbf{0.827±0.027}    &\textbf{0.842±0.026}  &0.838±0.027      &\textbf{0.849±0.017} &\textbf{0.853±0.014}      &\textbf{0.865±0.017} \\
\cline{2-9}

\rule{0pt}{10pt}
& Transformer    &0.749±0.025   &0.821±0.026 &0.825±0.027    &0.827±0.020  &0.837±0.018 &0.845±0.014      &0.861±0.023     \\
&  +WeightLoss     &0.752±0.025   &0.823±0.016   &0.825±0.019  &0.829±0.014     &0.839±0.015 &0.849±0.016      &0.859±0.014     \\
&  +SuperLoss      &0.763±0.020    &\textbf{0.826±0.026}    &0.830±0.026    &\textbf{0.838±0.024}  &0.844±0.017     &0.848±0.015     &0.859±0.016    \\
& +\mnamee   &0.765±0.025     &\textbf{0.826±0.023}  &0.831±0.025    &0.836±0.028   &0.846±0.015   &0.850±0.014   &\textbf{0.863±0.012}    \\
&\textbf{+\mnameee}  &\textbf{0.766±0.025}     &0.825±0.025  &\textbf{0.834±0.026}    &0.837±0.027   &\textbf{0.848±0.014}   &\textbf{0.852±0.015}   &0.862±0.014\\
\bottomrule[0.8pt]
\end{tabular}}
\end{table*}

\begin{table*}[!t]
\caption{Task \ref{task:3} Continuous Classification Performance on Solving Forgetting and Overfitting (BWT$\uparrow$ and FWT$\uparrow$) of Different Losses}\label{tb:BWTFWT}
\resizebox{\textwidth}{!}{
\begin{tabular}{l|ccccc|ccccc}
\toprule[0.8pt]
   &CE  &WeightLoss  &SuperLoss    &\textbf{\mnamee}  &\textbf{\mnameee}    &CE  &WeightLoss  &SuperLoss    &\textbf{\mnamee}  &\textbf{\mnameee}  \\
\midrule[0.8pt]
UCR-EQ   &--0.153   &+0.104    &+0.043    &\textbf{+0.125}  &+0.115 &+0.217  &+0.287    &+0.246   &+0.364   &\textbf{+0.369}    \\
USHCN      &--0.092    &+0.071   &+0.019  &+0.081 &\textbf{+0.092} &+0.316    &+0.322   &+0.331  &+0.348 &\textbf{+0.350} \\
COVID-19      &--0.066 &+0.017    &+0.006    &+0.032  &\textbf{+0.039}     &+0.237  &+0.265  &+0.323  &\textbf{+0.415}  &+0.412\\
SEPSIS   &--0.015  &+0.012    &+0.014    &\textbf{+0.021} &+0.018 &+0.289   &+0.478   &+0.421    &+0.498 &\textbf{+0.510} \\
\bottomrule[0.8pt]
\end{tabular}}
\end{table*}

\begin{figure*}[t]
\centering
\includegraphics[width=0.92\linewidth]{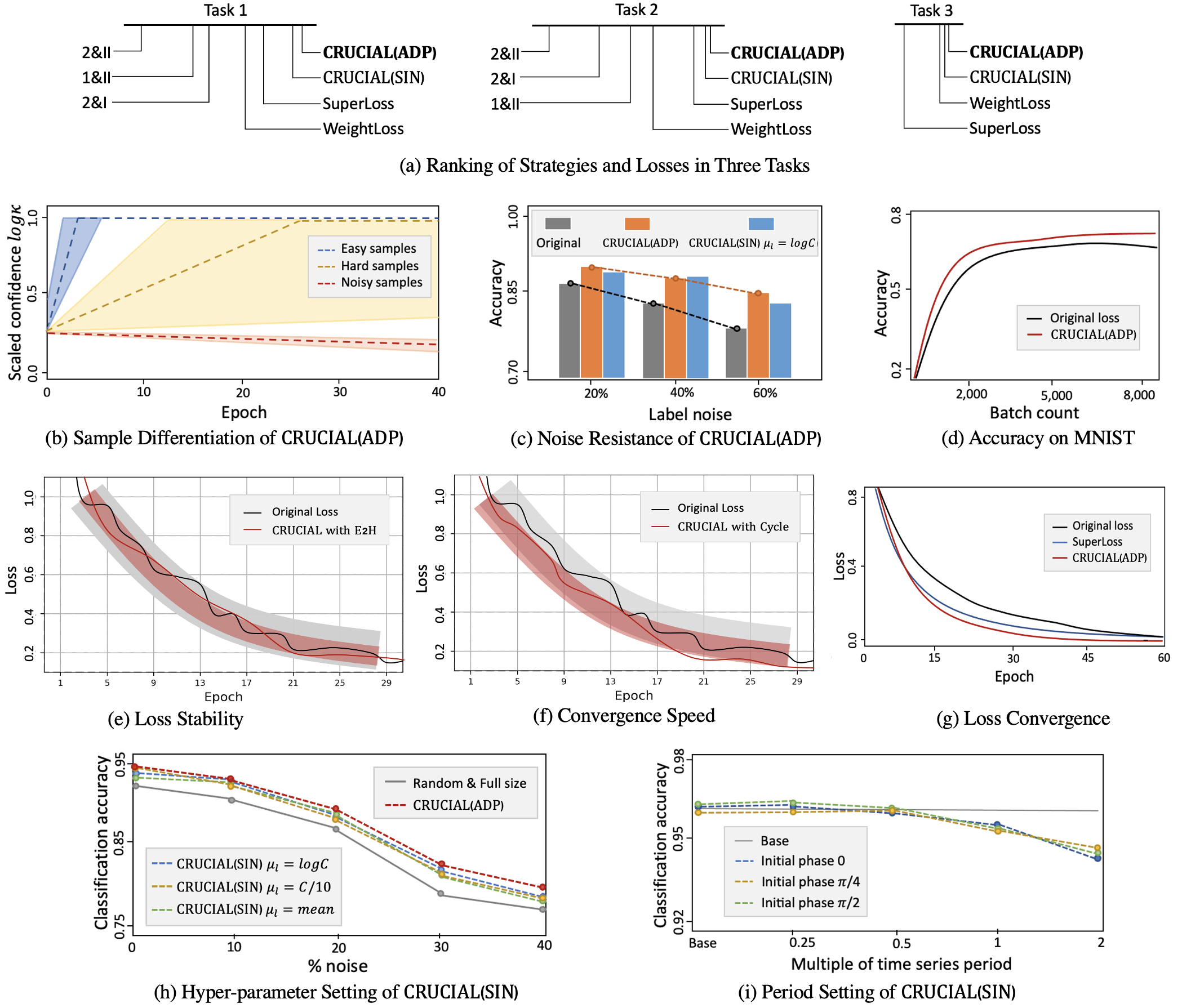}
\caption{{Performance of \mname with Different Settings}}
\label{fig:analysis}
\end{figure*}

\subsection{Results of Different Tasks} \label{sec:results}

\mnameee is significantly better than baselines in Bonferroni-Dunn tests as shown in Figure \ref{fig:analysis} (a): For task 1 $\text{Rank}_{\text{task1}}= 4.5>2.31+1 (k=7, n=2, m=5)$; For task 2, $\text{Rank}_{\text{task2}}= 4.5>1.34+1 (k=7, n=6, m=5)$; For task 3, $\text{Rank}_{\text{task3}}= 2.5>1.29+1 (k=4, n=2, m=5)$. $k, n, m$ are the number of methods, datasets, cross-validation fold, $CD=q_{\alpha}\sqrt{\frac{k(k+1)}{6(nm)}}$, if the average rank of baselines $> CD+1$, the result is significantly improved.

\subsubsection{Results of task 1}
As shown in Table \ref{tb:regression results}, \mname can improve the regression accuracy of base models and performs better than the compared strategies and losses. It also shows that the two strategies, easy-to-hard order (2) and increasing size (II), can improve the results respectively, but their combination will reduce the results. The easy-to-hard order is not suitable to be combined with the increasing size, but can be combined with the cyclical size, which shows the effectiveness of \mname.

\subsubsection{Results of task 2}
As shown in Table \ref{tb:classification accuracy 2}, for the LSTM model, \mname loss is better than all of the other losses, showing its additional properties, dynamic easy-to-hard learning order, and cyclical sample size, which can enhance the ability of confidence-aware losses to make deep learning models more accurate. Meanwhile, in most cases, especially for early classification tasks, \mnameee is better than \mnamee, demonstrating the advantages of the adaptive cycle.

\subsubsection{Results of task 3}
As shown in Table \ref{tb:classification accuracy 1}, \mnameee can improve the performance of DL models, making them classify more accurately at every time. Take sepsis diagnosis as an example, compared with the best baseline, our method improves the accuracy by 1.32\% on average, 2.19\% in the early 50\% time stage when the key features are unobvious. Each hour of delayed treatment increases sepsis mortality by 4-8\% \cite{seymour2017time}. With the same accuracy, we can predict 0.944 h in advance.

As shown in Table \ref{tb:BWTFWT}, \mnameee has the highest BWT, meaning it has the lowest negative influence that learning the new tasks has on the old tasks. \mnameee has the highest FWT, meaning it has the highest positive influence that learning the former data distributions has on the tasks. Our method can avoid model overfitting and guarantee certain model generalization.

\subsection{Discussions about Characteristics of \mname} \label{sec:analysis}
\subsubsection{\mname can schedule an easy-to-hard curriculum by dynamically modulating the loss amplitude}

\mname can schedule an easy-to-hard order without extra procedure when training the model. This mechanism is achieved by changing samples' losses. As shown in Figure \ref{fig:analysis} (b), it amplifies the contribution of easy samples while strongly flattening the input loss for hard ones: \mname naturally induces a delay (i.e. the time of convergence) with potential inconsistencies between the true status of a sample and its respective confidence.

Meanwhile, \mname has certain anti-noise potential. Time series dataset often has noise due to various problems in the process of data collection. Especially in medical scenarios, failure of monitoring equipment and manual error could introduce noise. The noise sample tends to get a large loss. If we scale the loss values of noise samples, the model could be resistant to noise. \mnameee has certain anti-noise because it can reduce the influence of noise on the model as shown in Figure \ref{fig:analysis} (c). The reason for this benefit is that \mnameee can recognize noise. As shown in Figure \ref{fig:analysis} (b), noisy samples (i.e. with wrong labels) tend to be harder and receive smaller weights throughout training. This side effect makes our method attractive when clean data are expensive while noisy data are widely available.

\subsubsection{\mnameee can adjust the size of training set adaptively with a rough cycle when training the model}

\mname achieves less error through the cyclical loss distribution as shown in Figure \ref{fig:cycle} (f). 

{\mnameee can adaptively find the suitable cycle for different datasets and tasks. As shown in Figure \ref{fig:cycle} (b) (c), for one-shot classification tasks, cycles for the COVID-19 dataset and SEPSIS dataset in \mnameee are roughly the same as cycles of 4 epochs and 16 epochs in \mnameee. More carefully, the cycle is mainly due to the joint action of dataset size and loss distribution \textsc{SK}. As shown in Figure \ref{fig:cycle} (d) (e), these two indicators are consistent with a certain phase difference. And the size is generally periodic and increasing.}

\mnamee is less sensitive to $\mu_{l}$. As shown in Figure \ref{fig:analysis} (h), for the sepsis classification task, we set $\mu_{l}$ to $\log2, 0.2, mean$. $\log C$ ($C$ is the number of classes) appears to be a natural boundary in the cross-entropy loss. $0.2$ is the empirical value, and $mean$ is the average loss in each epoch. Their results are similar. We suggest setting $\mu_{l}$ to $\log C$ for classification tasks and set $\mu_{l}$ to empirical value for regression tasks; \mnamee has different performance when using the different periodic function $\mathcal{F}$. The period of \mnamee should preferably be less than that of samples as shown in Figure \ref{fig:analysis} (i). According to experiments, we suggest setting the period of the method to be less than the period of the data, and their periods cannot be aligned. Thus, we give a reference setting: $\mathcal{F}=\sin^{2}{(\omega t+\mathcal{T})}$, $\omega=\frac{1}{3}\omega_{data}$ and $\mathcal{T}=0$.

\subsubsection{Combining two characteristics can promote the effect of model convergence}

\mname helps to make the model converge faster and better as shown in Figure \ref{fig:analysis} (g), the model with \mnameee converges faster and achieves a lower final error value. For continuous classification, as shown in Figure \ref{fig:method} right. The easy-to-hard order reduces the turbulence of the gradient. The use of partial harder samples avoids the overfitting problem. And the combination of using partial samples and starting from easy samples avoids the forgetting problem.

More carefully, the easy-to-hard mechanism (E2H) makes the convergence process more stable: As shown in Figure \ref{fig:analysis} (e), the curve of the loss value becomes smoother after using E2H, which is the reflection of the black dotted route of the solution for distribution 1 turning into the red route in Figure \ref{fig:method} right. The cycle size mechanism makes the convergence process faster: As shown in Figure \ref{fig:analysis} (f), the global error becomes smaller at most epochs after using the cycle size, which is the reflection of the black dotted route with overfitting and forgetting turning into the red route.

\section{Conclusion and Future Work}
This paper proposes a novel \mname loss to improve the performance of deep learning models when learning time series. It is task-agnostic and can be plugged on top of the original task loss with no extra procedure. We show that \mname can determine the sample contribution dynamically by applying the core principle of easy-to-hard order in curriculum learning; We prove that the cyclically changed set size can make expected errors between the updated gradient and optimal gradient smaller lower than the monotonous set size. \mnameee has the ability to create an adaptive cycle, which is more effective than setting the cycle manually \mnamee. According to the experiments, \mname has the best result than baselines: It helps the deep learning model to archive higher accuracy and faster convergence with certain anti-noise abilities. 

{Future work can apply \mname in other scenarios such as computer vision and natural language processing. The core idea of sample differentiation in \mname has the potential to be extended to other data forms like image data. We use fashion MNIST dataset \cite{DBLP:journals/corr/abs-1708-07747} to make a preliminary attempt and find that the convergence speed becomes faster and the accuracy becomes higher as shown in Figure \ref{fig:analysis} (d). Meanwhile, a category of research that evaluates samples, rankings, models, tasks, etc through difficulty, confidence, loss, etc has the fundamental idea of discrimination. They are interlinked to some extent. Thus, a unified framework for measurement can be studied in future work. It also has the potential for self-supervised comparative learning in addition to classical supervised learning. }

\section*{Acknowledgment}

This work was supported by the National Natural Science Foundation of China (No.62172018, No.62102008).

\bibliographystyle{IEEEtran}
\bibliography{ref}

\begin{thebibliography}{10}
\providecommand{\url}[1]{#1}
\csname url@samestyle\endcsname
\providecommand{\newblock}{\relax}
\providecommand{\bibinfo}[2]{#2}
\providecommand{\BIBentrySTDinterwordspacing}{\spaceskip=0pt\relax}
\providecommand{\BIBentryALTinterwordstretchfactor}{4}
\providecommand{\BIBentryALTinterwordspacing}{\spaceskip=\fontdimen2\font plus
\BIBentryALTinterwordstretchfactor\fontdimen3\font minus \fontdimen4\font\relax}
\providecommand{\BIBforeignlanguage}[2]{{%
\expandafter\ifx\csname l@#1\endcsname\relax
\typeout{** WARNING: IEEEtran.bst: No hyphenation pattern has been}%
\typeout{** loaded for the language `#1'. Using the pattern for}%
\typeout{** the default language instead.}%
\else
\language=\csname l@#1\endcsname
\fi
#2}}
\providecommand{\BIBdecl}{\relax}
\BIBdecl

\bibitem{9210118}
Q.~Ma, S.~Li, and G.~W. Cottrell, ``Adversarial joint-learning recurrent neural network for incomplete time series classification,'' \emph{IEEE Transactions on Pattern Analysis and Machine Intelligence}, vol.~44, no.~4, pp. 1765--1776, 2022.

\bibitem{9721108}
V.~Le~Guen and N.~Thome, ``Deep time series forecasting with shape and temporal criteria,'' \emph{IEEE Transactions on Pattern Analysis and Machine Intelligence}, pp. 1--1, 2022.

\bibitem{Sun_2022}
C.~Sun, S.~Hong, J.~Wang, X.~Dong, F.~Han, and H.~Li, ``A systematic review of deep learning methods for modeling electrocardiograms during sleep,'' \emph{Physiological Measurement}, vol.~43, no.~8, p. 08TR02, 2022.

\bibitem{9416768}
G.~Spadon, S.~Hong, B.~Brandoli, S.~Matwin, J.~F. Rodrigues-Jr, and J.~Sun, ``Pay attention to evolution: Time series forecasting with deep graph-evolution learning,'' \emph{IEEE Transactions on Pattern Analysis and Machine Intelligence}, vol.~44, no.~9, pp. 5368--5384, 2022.

\bibitem{DBLP:conf/cikm/SunSCZH022}
C.~Sun, M.~Song, D.~Cai, B.~Zhang, S.~Hong, and H.~Li, ``Confidence-guided learning process for continuous classification of time series,'' in \emph{Proceedings of the 31st {ACM} International Conference on Information {\&} Knowledge Management}, M.~A. Hasan and L.~Xiong, Eds., 2022, pp. 4525--4529.

\bibitem{DBLP:conf/icml/HacohenW19}
G.~Hacohen and D.~Weinshall, ``On the power of curriculum learning in training deep networks,'' in \emph{Proceedings of International Conference on Machine Learning (ICML)}, vol.~97, 2019, pp. 2535--2544.

\bibitem{DBLP:journals/corr/abs-2010-12493}
C.~Sun, S.~Hong, M.~Song, and H.~Li, ``A review of deep learning methods for irregularly sampled medical time series data,'' \emph{CoRR}, vol. abs/2010.12493, 2020.

\bibitem{DBLP:conf/icml/ArazoOAOM19}
E.~Arazo, D.~Ortego, P.~Albert, N.~E. O'Connor, and K.~McGuinness, ``Unsupervised label noise modeling and loss correction,'' in \emph{Proceedings of the 36th International Conference on Machine Learning, {ICML} 2019, 9-15 June 2019, Long Beach, California, {USA}}, ser. Proceedings of Machine Learning Research, vol.~97.\hskip 1em plus 0.5em minus 0.4em\relax {PMLR}, 2019, pp. 312--321.

\bibitem{DBLP:conf/ijcai/SunHSCSC021}
C.~Sun, S.~Hong, M.~Song, Y.~Chou, Y.~Sun, D.~Cai, and H.~Li, ``{TE-ESN:} time encoding echo state network for prediction based on irregularly sampled time series data,'' in \emph{Proceedings of the Thirtieth International Joint Conference on Artificial Intelligence}, 2021, pp. 3010--3016.

\bibitem{DBLP:conf/iclr/WuDN21}
X.~Wu, E.~Dyer, and B.~Neyshabur, ``When do curricula work?'' in \emph{9th International Conference on Learning Representations, {ICLR} 2021, Virtual Event, Austria, May 3-7, 2021}.\hskip 1em plus 0.5em minus 0.4em\relax OpenReview.net, 2021.

\bibitem{9349197}
M.~Delange, R.~Aljundi, M.~Masana, S.~Parisot, X.~Jia, A.~Leonardis, G.~Slabaugh, and T.~Tuytelaars, ``A continual learning survey: Defying forgetting in classification tasks,'' \emph{IEEE Transactions on Pattern Analysis and Machine Intelligence}, no.~7, pp. 3366--3385, 2022.

\bibitem{2014The}
W.~Chen, J.~Wang, Q.~L. Fe~Ng, S.~C. Xu, and L.~Ba, ``The treatment of severe and multiple injuries in intensive care unit: report of 80 cases,'' \emph{European Review for Medical \& Pharmacological Sciences}, vol.~18, no.~24, p. 3797, 2014.

\bibitem{sunpatterns}
C.~Sun, H.~Li, M.~Song, D.~Cai, B.~Zhang, and S.~Hong, ``Continuous diagnosis and prognosis by controlling the update process of deep neural networks,'' \emph{Patterns}, vol.~4, no.~2, p. 100687, 2023.

\bibitem{2021A}
D.~Kiyasseh, T.~Zhu, and D.~Clifton, ``A clinical deep learning framework for continually learning from cardiac signals across diseases, time, modalities, and institutions,'' \emph{Nature Communications}, vol.~12, no.~1, p. 4221, 2021.

\bibitem{DBLP:conf/cvpr/NovotnyALV18}
D.~Novotn{\'{y}}, S.~Albanie, D.~Larlus, and A.~Vedaldi, ``Self-supervised learning of geometrically stable features through probabilistic introspection,'' in \emph{2018 {IEEE} Conference on Computer Vision and Pattern Recognition, {CVPR} 2018, Salt Lake City, UT, USA, June 18-22, 2018}.\hskip 1em plus 0.5em minus 0.4em\relax Computer Vision Foundation / {IEEE} Computer Society, 2018, pp. 3637--3645.

\bibitem{DBLP:conf/cvpr/KendallGC18}
A.~Kendall, Y.~Gal, and R.~Cipolla, ``Multi-task learning using uncertainty to weigh losses for scene geometry and semantics,'' in \emph{2018 {IEEE} Conference on Computer Vision and Pattern Recognition, {CVPR} 2018, Salt Lake City, UT, USA, June 18-22, 2018}.\hskip 1em plus 0.5em minus 0.4em\relax Computer Vision Foundation / {IEEE} Computer Society, 2018, pp. 7482--7491.

\bibitem{DBLP:conf/nips/RevaudSHW19}
J.~Revaud, C.~R. de~Souza, M.~Humenberger, and P.~Weinzaepfel, ``{R2D2:} reliable and repeatable detector and descriptor,'' in \emph{Advances in Neural Information Processing Systems 32: Annual Conference on Neural Information Processing Systems 2019, NeurIPS 2019, December 8-14, 2019, Vancouver, BC, Canada}, H.~M. Wallach, H.~Larochelle, A.~Beygelzimer, F.~d'Alch{\'{e}}{-}Buc, E.~B. Fox, and R.~Garnett, Eds., 2019, pp. 12\,405--12\,415.

\bibitem{DBLP:conf/nips/SaxenaTD19}
S.~Saxena, O.~Tuzel, and D.~DeCoste, ``Data parameters: {A} new family of parameters for learning a differentiable curriculum,'' in \emph{Advances in Neural Information Processing Systems 32: Annual Conference on Neural Information Processing Systems 2019, NeurIPS 2019, December 8-14, 2019, Vancouver, BC, Canada}, H.~M. Wallach, H.~Larochelle, A.~Beygelzimer, F.~d'Alch{\'{e}}{-}Buc, E.~B. Fox, and R.~Garnett, Eds., 2019, pp. 11\,093--11\,103.

\bibitem{DBLP:conf/nips/CastellsWR20}
T.~Castells, P.~Weinzaepfel, and J.~Revaud, ``Superloss: {A} generic loss for robust curriculum learning,'' in \emph{Advances in Neural Information Processing Systems 33: Annual Conference on Neural Information Processing Systems 2020, NeurIPS 2020, December 6-12, 2020, virtual}, H.~Larochelle, M.~Ranzato, R.~Hadsell, M.~Balcan, and H.~Lin, Eds., 2020.

\bibitem{DBLP:journals/pami/WangCZ22}
X.~Wang, Y.~Chen, and W.~Zhu, ``A survey on curriculum learning,'' \emph{{IEEE} Trans. Pattern Anal. Mach. Intell.}, vol.~44, no.~9, pp. 4555--4576, 2022.

\bibitem{DBLP:conf/acl/ZhouYWWC20}
Y.~Zhou, B.~Yang, D.~F. Wong, and Y.~Wan, ``Uncertainty-aware curriculum learning for neural machine translation,'' in \emph{Proceedings of the Association for Computational Linguistics (ACL)}, 2020, pp. 6934--6944.

\bibitem{DBLP:conf/nips/KumarPK10}
M.~P. Kumar, B.~Packer, and D.~Koller, ``Self-paced learning for latent variable models,'' in \emph{Advances in Neural Information Processing Systems}, J.~D. Lafferty, C.~K.~I. Williams, J.~Shawe{-}Taylor, R.~S. Zemel, and A.~Culotta, Eds.\hskip 1em plus 0.5em minus 0.4em\relax Curran Associates, Inc., 2010, pp. 1189--1197.

\bibitem{DBLP:conf/nips/ZhouWB20}
T.~Zhou, S.~Wang, and J.~A. Bilmes, ``Curriculum learning by dynamic instance hardness,'' in \emph{Advances in Neural Information Processing Systems}, H.~Larochelle, M.~Ranzato, R.~Hadsell, M.~Balcan, and H.~Lin, Eds., 2020.

\bibitem{DBLP:conf/icml/XuT22}
Z.~Xu and A.~Tewari, ``On the statistical benefits of curriculum learning,'' in \emph{International Conference on Machine Learning, {ICML} 2022}, vol. 162, 2022, pp. 24\,663--24\,682.

\bibitem{DBLP:conf/nips/0003GY20}
Y.~Liu, Y.~Gao, and W.~Yin, ``An improved analysis of stochastic gradient descent with momentum,'' in \emph{Advances in Neural Information Processing Systems 33: Annual Conference on Neural Information Processing Systems 2020, NeurIPS 2020, December 6-12, 2020, virtual}, H.~Larochelle, M.~Ranzato, R.~Hadsell, M.~Balcan, and H.~Lin, Eds., 2020.

\bibitem{DBLP:conf/icml/Axelrod0SV20}
B.~Axelrod, S.~Garg, V.~Sharan, and G.~Valiant, ``Sample amplification: Increasing dataset size even when learning is impossible,'' in \emph{Proceedings of the 37th International Conference on Machine Learning, {ICML} 2020}, vol. 119, 2020, pp. 442--451.

\bibitem{sunapin}
C.~Sun, H.~Li, M.~Song, D.~Cai, B.~Zhang, and S.~Hong, ``Adaptive model training strategy for continuous classification of time series,'' \emph{Applied Intelligence}, 2023.

\bibitem{DBLP:journals/corr/abs-2202-05531}
\BIBentryALTinterwordspacing
H.~T. Kesgin and M.~F. Amasyali, ``Cyclical curriculum learning,'' \emph{CoRR}, vol. abs/2202.05531, 2022. [Online]. Available: \url{https://arxiv.org/abs/2202.05531}
\BIBentrySTDinterwordspacing

\bibitem{SILSO}
S.~W.~D. Center, ``The international sunspot number, int. sunspot number monthly bull. online catalogue (1749-2016),'' \url{http://www.sidc.be/silso/}, 2016.

\bibitem{USHCN}
W.~C. Menne, M. and V.~R., ``Long-term daily and monthly climate records from stations across the contiguous united states.'' \emph{[Online]}, 2010.

\bibitem{UCRArchive}
Y.~Chen, E.~Keogh, B.~Hu, N.~Begum, A.~Bagnall, A.~Mueen, and G.~Batista, ``The ucr time series classification archive,'' July 2015, \url{www.cs.ucr.edu/~eamonn/time_series_data/}.

\bibitem{DBLP:conf/cinc/ReynaJSJSWSNC19}
\BIBentryALTinterwordspacing
M.~A. Reyna, C.~Josef, S.~Seyedi, R.~Jeter, S.~P. Shashikumar, M.~B. Westover, A.~Sharma, S.~Nemati, and G.~D. Clifford, ``Early prediction of sepsis from clinical data: the physionet/computing in cardiology challenge 2019,'' in \emph{46th Computing in Cardiology, CinC 2019, Singapore, September 8-11, 2019}.\hskip 1em plus 0.5em minus 0.4em\relax {IEEE}, 2019, pp. 1--4. [Online]. Available: \url{https://doi.org/10.23919/CinC49843.2019.9005736}
\BIBentrySTDinterwordspacing

\bibitem{COVID-19}
G.~J. e.~a. Yan~L, Zhang H~T, ``An interpretable mortality prediction model for covid-19 patients,'' \emph{Nature, Machine intelligence}, vol.~2, 2020.

\bibitem{DBLP:journals/BMC/sun}
C.~Sun, S.~Hong, M.~Song, H.~Li, and Z.~Wang, ``Predicting covid-19 disease progression and patient outcomes based on temporal deep learning,'' \emph{BMC Medical Informatics and Decision Making}, vol. 21:45, 2020.

\bibitem{DBLP:journals/tai/GuptaGBD20}
A.~Gupta, H.~P. Gupta, B.~Biswas, and T.~Dutta, ``Approaches and applications of early classification of time series: {A} review,'' \emph{{IEEE} Trans. Artif. Intell.}, vol.~1, no.~1, pp. 47--61, 2020.

\bibitem{Baytas:2017:PSV:3097983.3097997}
I.~M. Baytas, C.~Xiao, X.~Zhang, F.~Wang, A.~K. Jain, and J.~Zhou, ``Patient subtyping via time-aware lstm networks,'' in \emph{Proceedings of the 23rd ACM SIGKDD International Conference on Knowledge Discovery and Data Mining}, ser. KDD '17.\hskip 1em plus 0.5em minus 0.4em\relax ACM, 2017, pp. 65--74.

\bibitem{DBLP:journals/nn/0058S21}
W.~Chen and K.~Shi, ``Multi-scale attention convolutional neural network for time series classification,'' \emph{Neural Networks}, vol. 136, pp. 126--140, 2021.

\bibitem{DBLP:journals/datamine/FawazFWIM19}
H.~I. Fawaz, G.~Forestier, J.~Weber, L.~Idoumghar, and P.~Muller, ``Deep learning for time series classification: a review,'' \emph{Data Min. Knowl. Discov.}, vol.~33, no.~4, pp. 917--963, 2019.

\bibitem{DBLP:conf/aaai/ZhouZPZLXZ21}
H.~Zhou, S.~Zhang, J.~Peng, S.~Zhang, J.~Li, H.~Xiong, and W.~Zhang, ``Informer: Beyond efficient transformer for long sequence time-series forecasting,'' in \emph{Thirty-Fifth {AAAI} Conference on Artificial Intelligence, {AAAI} 2021, Thirty-Third Conference on Innovative Applications of Artificial Intelligence, {IAAI} 2021, The Eleventh Symposium on Educational Advances in Artificial Intelligence, {EAAI} 2021, Virtual Event, February 2-9, 2021}.\hskip 1em plus 0.5em minus 0.4em\relax {AAAI} Press, 2021, pp. 11\,106--11\,115.

\bibitem{DBLP:conf/kdd/ZerveasJPBE21}
\BIBentryALTinterwordspacing
G.~Zerveas, S.~Jayaraman, D.~Patel, A.~Bhamidipaty, and C.~Eickhoff, ``A transformer-based framework for multivariate time series representation learning,'' in \emph{{KDD} '21: The 27th {ACM} {SIGKDD} Conference on Knowledge Discovery and Data Mining, Virtual Event, Singapore, August 14-18, 2021}, F.~Zhu, B.~C. Ooi, and C.~Miao, Eds.\hskip 1em plus 0.5em minus 0.4em\relax {ACM}, 2021, pp. 2114--2124. [Online]. Available: \url{https://doi.org/10.1145/3447548.3467401}
\BIBentrySTDinterwordspacing

\bibitem{DBLP:journals/ijon/RizkA19}
Y.~Rizk and M.~Awad, ``On extreme learning machines in sequential and time series prediction: {A} non-iterative and approximate training algorithm for recurrent neural networks,'' \emph{Neurocomputing}, vol. 325, pp. 1--19, 2019.

\bibitem{seymour2017time}
C.~W. Seymour, F.~Gesten, H.~C. Prescott, M.~E. Friedrich, T.~J. Iwashyna, G.~S. Phillips, S.~Lemeshow, T.~Osborn, K.~M. Terry, and M.~M. Levy, ``Time to treatment and mortality during mandated emergency care for sepsis,'' \emph{New England Journal of Medicine}, vol. 376, no.~23, pp. 2235--2244, 2017.

\bibitem{DBLP:journals/corr/abs-1708-07747}
H.~Xiao, K.~Rasul, and R.~Vollgraf, ``Fashion-mnist: a novel image dataset for benchmarking machine learning algorithms,'' \emph{CoRR}, vol. abs/1708.07747, 2017.

\end{thebibliography}

\clearpage

\appendix

\begin{center}
\Large \quad \\  \textbf{Appendix for\\ Curricular and Cyclical Loss for Time Series Learning Strategy}\\
\quad \\

\end{center}

\normalsize

\section{Expression of \mname} \label{app:expression}

\subsection{General Expression}

The general expression of \mname is:
\begin{equation} 
L=\kappa \cdot \varphi_{D}(l,\epsilon)+\varphi_{C}(\kappa), \quad \epsilon=\varphi_{S}(\varepsilon)\nonumber
\end{equation}

It has three key functions:
\begin{itemize}
    \item \textbf{Constraint function $\varphi_{C}$}
    \item \textbf{Differentiation function $\varphi_{D}$}
    \item \textbf{State function $\varphi_{S}$}
\end{itemize}

\subsection{Expression of \mnameee}

\mnamee is a explicit cyclical equation option for \mname. It is expressed as
\begin{equation} 
\begin{aligned}
    &\text{\mnamee}(l_{i,t})
    =\left\{ \begin{aligned} 
    \kappa_{i,t}\cdot(l_{i,t}-\epsilon_{t})&+\lambda_{t}\cdot(\log\kappa_{i,t})^{2} \ \ l_{i,t}\geq \varepsilon_{t}\\
    &0 \quad\quad\quad\quad\quad\quad\ \ \ l_{i,t}<\varepsilon_{t}
    \end{aligned} \right.\\ 
    &\text{where } \kappa_{i,t}=e^{-\frac{1}{2}W\max(-\frac{2}{e},\frac{l_{i,t}-\epsilon}{\lambda})}
\end{aligned} \nonumber
\end{equation}

Its three key functions are:
\begin{itemize}
    \item Constraint function $\varphi_{C}(\kappa)=\lambda\cdot(\log\kappa)^{2}$
    \item Differentiation function $\varphi_{D}(l,\epsilon)=l-\epsilon$
    \item State function $\varphi_{S}(\varepsilon)=2\varepsilon- \mu_{l}$
\end{itemize}

To achieve the cyclical size, we define $\varepsilon$ by the periodic function $\mathcal{F}$. $\varepsilon$, $\epsilon$ and $\lambda$ are in the same rhythm. $\mathcal{F}$ is optional and we found that $\mathcal{F}=\sin^{2}{(\omega t+\mathcal{T})}, t\in \mathbb{N}$ performs well in experiments:
\begin{itemize}
    \item $\mathcal{F}=\sin^{2}{(\omega t+\mathcal{T})}, t\in \mathbb{N}$ 
    \item $\lambda=-\log{\mathcal{F}}$
    \item $\varepsilon=\frac{1}{2}\mathcal{F}\cdot \mu_{l}$  
\end{itemize}

\begin{proof}[\mnamee is periodic] \
\mnamee becomes equivalent to the original input loss when $\lambda$ tends to infinity:\\
As 
\begin{equation}\nonumber
     \lim\limits_{\lambda\rightarrow+\infty}\kappa_{i}=e^{-\frac{W}{2}\max(-\frac{2}{e},\frac{l_{i}-\epsilon}{\lambda})}=e^{-0}
\end{equation}
Hence 
\begin{equation}\nonumber
     \lim\limits_{\lambda\rightarrow+\infty}\text{\mnamee}_{\lambda}(l_{i})=\lim\limits_{\lambda\rightarrow+\infty}l_{i}-\epsilon+\lambda W(\frac{l_{i}-\epsilon}{2\lambda})^2=l_{i}-\epsilon
\end{equation}
The loss fails when $\lambda$ tends to $0$:\\
As 
\begin{equation}\nonumber
     \lim\limits_{\lambda\rightarrow0}\kappa _{i}=e^{-\infty}=0
\end{equation}
Hence 
\begin{equation}\nonumber
\lim\limits_{\lambda\rightarrow+\infty}\text{\mnamee}_{\lambda}(l_{i})=0
\end{equation}
\text{\mname} is periodic and the length of the period is 4 epochs ($n\in \mathbb{N}$):
\begin{equation}\nonumber
\begin{aligned}
&\text{\mname}_{t=4n}(l_{i})=l_{i}-\mu_{l}, l_{i}\geq 0;\\
&\text{\mname}_{t=4n+1\lor 4n+3}(l_{i})=\kappa (l_{i}-\frac{\mu_{l}}{2})+0.1(\log\kappa)^{2}, \\
&\quad \quad \quad \quad \kappa =\left\{ \begin{aligned} 
    &e^{-\frac{1}{2}W(10l_{i}-5\mu_{l})},l_{i}\geq\frac{\mu_{l}}{2}-\frac{1}{5e}\\
    &e^{-\frac{1}{2}W(-\frac{2}{e})},\frac{\mu_{l}}{2}-\frac{1}{5e}>l_{i}\geq \frac{1}{4}\mu_{l}
\end{aligned} \right.;\\
&\text{\mname}_{4n+2}(l_{i})=0,l_{i}\geq \frac{1}{2}\mu_{l}.
\end{aligned}
\end{equation}
\end{proof}

\subsection{Expression of \mnameee}

\mnameee is an adaptive cyclical equation option for \mname. It is expressed as
\begin{equation} 
\begin{aligned}
    &\text{\mname}(l_{i,t})=\kappa_{i,t}\cdot(l_{i,t}-\epsilon_{t})+ \lambda\cdot(\log\kappa_{i,t})^{2}\\
    &\text{where } \kappa_{i,t}=e^{-\frac{1}{2}W\max(-\frac{2}{e},\frac{l_{i,t}-\epsilon_{t}}{\lambda})},\\
    &\text{and } \epsilon_{t}=\frac{\sum_{i}^{N}(l_{i,t-1}-\mu_{l_{t-1}})^{3}\cdot \mu_{l_{t-1}}}{N\sigma_{l_{t-1}}^{3}}
\end{aligned}\nonumber
\end{equation}

Its three key functions are:
\begin{itemize}
    \item Constraint function $\varphi_{C}(\kappa)=\lambda\cdot(\log\kappa)^{2}$
    \item Differentiation function $\varphi_{D}(l,\epsilon)=l-\epsilon$
    \item State function $\varphi_{S}(\varepsilon)=\textsc{sk}=\frac{\sum_{i}^{N}(l_{i}-\mu_{l})^{3}}{N\sigma_{l}^{3}}$
\end{itemize}

To define the confidence coefficient $\kappa$, we utilize the regularization term $\lambda(\log\kappa)^{2}$ for constraint function $\psi$. In this way, we can use the method of closed-form formula \cite{DBLP:conf/nips/CastellsWR20} to obtain the final value $\kappa^{*}=\arg\min_{\kappa}L(l,\kappa)$. Where we directly use their converged value at the limit, which only depends on the input loss $l$, without requiring additional parameters or inducing a convergence delay. 

\begin{proof}[The confidence parameters $\kappa$ does not need to be learned and are up-to-date with the sample status under the operation of constraint function $\varphi_{C}$] 
With the exponential mapping 
\begin{equation}\nonumber
    \kappa=e^{x}
\end{equation}
and scaled mapping 
\begin{equation}\nonumber
    \beta=\frac{l-\epsilon}{\lambda}
\end{equation}
loss equation can be rewritten as 
\begin{equation}\nonumber
    \min L=\min_{\kappa}\kappa(l-\epsilon)+\lambda(\log\kappa)^{2}=\min_{x}\beta e^{x}+x^{2}
\end{equation} 
When 
\begin{equation}\nonumber
    \beta>-\frac{2}{e}
\end{equation} 
the function admits a single local minimum by solving the root of the derivative
\begin{equation}\nonumber
    \frac{\partial(\beta e^{x}+x^{2})}{\partial x}=0\Rightarrow \frac{\beta}{2}=-xe^{-x}\Rightarrow x=-W(\frac{\beta}{2})
\end{equation} 
where $W(*)$ is the Lambert W function for the solution to $y=xe^{x}$. \\
Finally, when $\beta>-\frac{2}{e}$,
\begin{equation}\nonumber
    \kappa=e^{-W(\frac{\beta}{2})}
\end{equation} 
When $\beta\leq -\frac{2}{e}$,
we cap the optimal 
\begin{equation}\nonumber
    \kappa=e^{-W(\frac{-2/e}{2})}=e
\end{equation} 
\end{proof}
Note that $\kappa_{i,t}=e^{-\frac{1}{2}W\max(-\frac{2}{e},\frac{l_{i,t}-\epsilon_{t}}{\lambda})}$ applies to both \mnamee and \mnameee.

\section{Proofs of properties and theorems} \label{app:proof}
\mname satisfy three basic properties and two characteristic properties. 

\subsection{Basic Properties}

\mname satisfy three basic properties for generality. 

\vspace{+0.2cm}
\noindent \rm{\textbf{Property 1 Translation-invariance.}} $\forall C, \exists C'|L(l+C,\kappa)=L(l,\kappa)+C'$. $C$, $C'$ are constant.

\vspace{+0.2cm}
\noindent\rm{\textbf{Property 2 Homogeneity.}} $\exists \lambda, \lambda'|\forall C>0, L_{\lambda'}(C\cdot l,\kappa)=C\cdot L_{\lambda}(l,\kappa)$. $C$ is a constant.

\vspace{+0.2cm}
\noindent\rm{\textbf{Property 3 Generalization.}} $\exists \kappa | L(l,\kappa)=l+C$. $C$ is a constant.

\vspace{+0.2cm}
\begin{proof}[Property \ref{prop:1}] 
\begin{equation}\nonumber
    \text{\mname}(l+C)=\text{\mname}(l)+C', C'=(\kappa'-\kappa)(l-\epsilon)+\kappa'C+\lambda((\log\kappa')^{2}-(\log\kappa)^{2})
\end{equation} 
\end{proof}

\begin{proof}[Property \ref{prop:2}] 
\begin{equation}\nonumber
    \lambda'=C\lambda, \text{\mname}_{C\lambda}(Cl)=\kappa(Cl)(Cl-C\epsilon)+C\lambda(\log\kappa(Cl))^{2}=C(\kappa (l)(l-\epsilon)+\lambda(\log\kappa (l))^{2})=C\cdot\text{\mname}_{\lambda}(l)
\end{equation} 
\end{proof}

\begin{proof}[Property \ref{prop:3}] 
\begin{equation}\nonumber
    \kappa=1,\text{\mname}(l)=1\cdot (l-\epsilon)+\lambda\cdot(\log1)^{2}=l-\epsilon
\end{equation} 
\end{proof}

With Property \ref{prop:1}, if we add a constant to the input, loss should have no effect on loss’s gradient; With Property \ref{prop:2}, we can accordingly rescale $\lambda$ to handle input losses of any amplitude; With Property \ref{prop:3}, it can amount to the input loss for a particular confidence.

\subsection{Characteristic Properties}

After satisfying the three basic properties, \mname satisfies two additional points defined in Definition \ref{def:ccl}. They are achieved by Property \ref{prop:4} and Property \ref{prop:5}, which are also the main points of our method.

\vspace{+0.2cm}
\noindent\rm{\textbf{Property 4 Differentiated Scaling.}}
$\forall l_{i},l_{j},l_{i}<\epsilon,l_{j}>\epsilon|\frac{L_{i}}{l_{i}-\epsilon}>\frac{L_{j}}{l_{j}-\epsilon}$

\vspace{+0.2cm}
\noindent\rm{\textbf{Property 5 Cyclical Sampling with Less Total Error.}}
$\exists\varepsilon, \forall C| \varepsilon_{t} \approx\varepsilon_{t+p}, \sum_{i=1}^{N}L_{\varepsilon}(l_{i})<\sum_{i=1}^{N}L_{C}(l_{i})$, $p$ is the period and $C$ is the constant threshold.

\vspace{+0.2cm}
When continuously learning time series over time, the model learns subsets $\mathcal{S}_{t}=\{X_{i,1:t}\}_{i=1}^{N}$ one by one from $\mathcal{S}$. Due to uncertain similarity and difference between adjacent $X_{1:t}$ and $X_{1:t-1}$, we arrange the size from non-full/partial dataset, specifically, the cyclical size. Property \ref{prop:5} shows that the cyclical size of the learning set can make the model achieve better performance. We prove that in the learning process, the expected error between the updated gradient and optimal gradient by using the alternate data size is smaller than that by using monotonous ones. The process is Theorem \ref{theo:1} $+$ Theorem \ref{theo:2} $\rightarrow$ Theorem \ref{theo:3}. 

\vspace{+0.2cm}
{Let the $\mathcal{L}$ function be minimized for the data set $\mathcal{S}=\{X_{i}\}_{i=1}^{N}$. $\mathcal{S}$ dataset consist of $N$ samples, where $X_{i}$ represents the $i$-th training sample and $y_i$ is its label. When training a deep learning model $f$ with parameters $w$, we minimize the loss $l_{i}$ of each input sample $X_{i}$ to optimize global loss $\mathcal{L}$:}
\begin{equation} \label{eq:objective}
    \arg \min \mathcal{L}_{w}
\end{equation}
\begin{equation} \label{eq:loss}
\begin{aligned}
    \mathcal{L}_{w}&=\frac{1}{N}\sum^{N}_{i=1}l_{i}\\
    l_{i}=&l(y_{i},f(X_{i},w)) 
\end{aligned}
\end{equation}
Where $f(X_{i}, w)$ is the prediction for $X_{i}$ of model $f$ with parameters $w$, $y_{i}$ is the actual label and $l$ is the loss function.

The optimization process for Equation \ref{eq:objective} can be performed with the gradient descent algorithm 
\begin{equation}
   w_{t+1}=w_{t}-\eta\nabla_{w}\mathcal{L}(w_{t}) 
\end{equation}
Where it optimizes global loss $\mathcal{L}$ with learning rate $\eta$. 

But in practice, the optimization is performed based on data batch, like (momentum) stochastic gradient descent:
\begin{equation}
   w_{t+1}=w_{t}-\eta\nabla_{w}l_{i}(w_{t})
\end{equation}

The difference of the stochastic gradient descent algorithm from the gradient descent algorithm is that it updates the weights at each step, not according to all samples in the training set, but according to a small number of randomly selected samples at each step. Thus, it makes some errors. 

The expected value of mean squared error between updating with sample $X_{i}$ and all samples at time $t$ is:
\begin{equation}
\begin{aligned}
    E[\Delta(w_{t})]&=E[(\nabla_{w}l_{i}(w_{t})-\nabla_{w}\mathcal{L}(w_{t}))^{2}]\\
    &=E[(\nabla_{w}(l_{i}(w_{t})-\mathcal{L}(w_{t})))^{2}]
\end{aligned}
\end{equation}

{Thus, at training time $t$, $w_{t+1}=w_{t}-\eta\nabla_{w}l_{i}(w_{t})$, the main factor leading to the difference $\delta_{t}$ between updating $w_{t}$ to $w_{t+1}$ and updating $w_{t}$ to optimal $\overline{w}_{t+1}$ is reflected in the mean squared error between the loss $l_{i}$ and the global loss $\mathcal{L}$.}
\begin{equation}\label{eq:error}
\begin{aligned}
    E[\delta_{t}]=E[(l_{i}(w_{t})-\mathcal{L}(w_{t}))^{2}]
\end{aligned}
\end{equation}

Based on this we focus on two conditions defined as below. 
\begin{equation}
\begin{aligned}
\textsc{Condition U: } &\text{Select samples randomly} \Rightarrow X_{i} \text{ is selected with the probability } \frac{1}{N}\\
\textsc{Condition P: } &\text{Select samples based on loss } l_{i} \Rightarrow X_{i} \text{ is selected with the probability }\frac{\Lambda e^{-\Lambda l_{i}}}{\sum_{i=1}^{N}\Lambda e^{-\Lambda l_{i}}}
\end{aligned}\nonumber
\end{equation}

\textsc{Condition U}: If the training samples are randomly selected from a uniform distribution, the $i$-th sample will be selected with the mean probability in Equation \ref{eq:U}, where $N$ is the number of samples.
\begin{equation}\label{eq:U}
    P_{i}=\frac{1}{N}
\end{equation}

\textsc{Condition P}: If the training samples are selected differently, the model training process will be the exponential distributed stochastic gradient descent. For example, in \mnameee, samples whose $l_{i}<\epsilon$ are more likely to be selected as their loss amplitudes are maintained but that of samples whose $l_{i}>\epsilon$ are modulated smaller. As the losses are uniform distributed, the probability $P(l_{i}<\epsilon)$ of a sample is exponential distributed with $p(l_{i})=\Lambda e^{-\Lambda l_{i}}$. $\Lambda$ is the rate parameter. The $i$-th sample will be selected with the exponential probability in Equation \ref{eq:P}.
\begin{equation}\label{eq:P}
    P_{i}=\frac{\Lambda e^{-\Lambda l_{i}}}{\sum_{i=1}^{N}\Lambda e^{-\Lambda l_{i}}}
\end{equation}

Based on above two conditions, we propose three theorems. Theorem \ref{theo:1} and \ref{theo:2} are the foundation of Theorem \ref{theo:3}.

\vspace{+0.2cm}
\noindent\rm{\textbf{Theorem 1}}
At training time $t$, if the loss values are normal distributed $\mathcal{N}(\mu,\sigma^{2})$, the expected error of U will be lower than that of P:
\begin{equation}
    l_{i}(w_{t})\sim \mathcal{N}(\mu,\sigma^{2})\Longrightarrow E_{U}<E_{P} \nonumber
\end{equation}

\noindent\rm{\textbf{Theorem 2}}
At training time $t$, if the loss values are half-normal distributed $\mathcal{N}(\mu+\sigma\sqrt{\frac{2}{\pi}},\sigma^{2}(1-\frac{2}{\pi}))$, the expected error of P will be lower than that of U under the condition of $\sigma<\frac{\pi}{\Lambda}$:
\begin{equation}
    l_{i}(w_{t})\sim \mathcal{N}(\mu+\sigma\sqrt{\frac{2}{\pi}},\sigma^{2}(1-\frac{2}{\pi}))\land\sigma<\frac{\pi}{\Lambda}\Longrightarrow E_{P}<E_{U} \nonumber
\end{equation}

\noindent\rm{\textbf{Theorem 3}}
The expected error of classical loss becomes lower with the constraint of \mname.
\begin{equation}
    E_{\text{\mname}}<E \nonumber
\end{equation}

At training time $t$, based on Equation \ref{eq:error} we write expected value of the error for U and P as follow using variance bias decomposition.
\begin{equation}\label{eq:EU}
    E_{U}[\delta_{t}]=(E_{U}[A]-B)^{2}+E_{U}[A^{2}]-E^{2}_{U}[A]
\end{equation}
\begin{equation}\label{eq:EP}
    E_{P}[\delta_{t}]=(E_{P}[A]-B)^{2}+E_{P}[A^{2}]-E^{2}_{P}[A]
\end{equation}
with 
\begin{equation}
   { A=l_{i}(w_{t})}
\end{equation}
\begin{equation}
   { B=\mathcal{L}(w_{t})}
\end{equation}

\begin{proof} [Theorem \ref{theo:1}]\ At training time $t$, $l_{i}(w_{t})\sim \mathcal{N}(\mu,\sigma^{2})$, \\
From Equation \ref{eq:loss}
\begin{equation}\nonumber
    B=\mathcal{L}(w_{t})= \frac{1}{N}\sum^{N}_{i=1}l_{i}(w_{t})=\mu
\end{equation}
Equation \ref{eq:EU} and can be calculated as follows:\\
\text{\quad\quad}Since $l_{i}(w_{t})\sim \mathcal{N}(\mu,\sigma^{2})$, and the $i$-th sample is selected with mean probability in Equation \ref{eq:U}
\begin{equation}
    E_{U}[A]=E_{U}[l_{i}(w_{t})]=\mu
\end{equation}
\text{\quad\quad}From the definition of variance $D(x)=E(x-E(x))^{2}=E(x^{2})-E^{2}(x)$
\begin{equation}
    E_{U}[A^{2}]-E_{U}^{2}[A]=\sigma^2 
\end{equation}
\text{\quad\quad}Thus
\begin{equation}
    E_{U}=(\mu-\mu)^{2}+\sigma^2=\sigma^{2}
\end{equation}
Equation \ref{eq:EP} and can be calculated as follows:\\
\text{\quad\quad}Since the $i$-th sample is selected with the exponential probability in Equation \ref{eq:P}
\begin{equation}
      E_{P}[A]=\frac{E[l_{i}(w_{t}) \Lambda  e^{-\Lambda  l_{i}(w_{t})}]}{E[\Lambda e^{-\Lambda l_{i}(w_{t})}]}=\frac{\Lambda e^{\frac{\Lambda(\Lambda\sigma^{2}-2\mu)}{2}}(\mu-\Lambda\sigma^{2})}{\Lambda e^{\frac{\Lambda(\Lambda\sigma^{2}-2\mu)}{2}}}=
      \mu-\Lambda\sigma^2
\end{equation}
\begin{equation}      
       E_{P}[A^2]=\frac{E[l_{i}^{2}(w_{t})\Lambda e^{-\Lambda \nabla_{w_{t}}l_{i}(w_{t})}]}{E[\Lambda e^{-\Lambda l_{i}(w_{t})}]}=\mu^2+\sigma^2+\Lambda^{2}\sigma^4-2\Lambda\mu\sigma^2  
\end{equation}
\text{\quad\quad}Thus 
\begin{equation}
    E_{P}=((\mu-\Lambda \sigma^2)-\mu)^{2}+\mu^2+\sigma^2+\Lambda^{2}\sigma^4-2\Lambda\mu\sigma^2-(\mu-\Lambda\sigma^2)^2=\Lambda^{2}\sigma^4+\sigma^2
\end{equation}
Finally, we compare $E_{U}$ and $E_{P}$
\begin{equation}\nonumber
    \sigma^2<\Lambda^{2}\sigma^4+\sigma^2 \Longrightarrow E_{U}<E_{P}
\end{equation}
\end{proof}

\begin{proof} [Theorem \ref{theo:2}]\ At training time $t$, $l_{i}(w_{t})\sim \mathcal{N}(\mu+\sigma\sqrt{\frac{2}{\pi}},\sigma^{2}(1-\frac{2}{\pi}))$, \\
From Equation \ref{eq:loss}
\begin{equation}\nonumber
    B=\mathcal{L}(w_{t})= \frac{1}{N}\sum^{N}_{i=1}l_{i}(w_{t})=\mu+\sigma\sqrt{\frac{2}{\pi}}
\end{equation}
Equation \ref{eq:EU} and can be calculated as follows:\\
\text{\quad\quad}Since $l_{i}(w_{t})\sim \mathcal{N}(\mu+\sigma\sqrt{\frac{2}{\pi}},\sigma^{2}(1-\frac{2}{\pi}))$, and the $i$-th sample is selected with mean probability in Equation \ref{eq:U}
\begin{equation}
     E_{U}[A]=\mu+\sigma\sqrt{\frac{2}{\pi}}
\end{equation}
\text{\quad\quad}From the definition of variance $D(x)=E(x-E(x))^{2}=E(x^{2})-E^{2}(x)$
\begin{equation}
     E_{U}[A^{2}]-E_{U}^{2}[A]=\sigma^{2}(1-\frac{2}{\pi})
\end{equation}
\text{\quad\quad}Thus
\begin{equation}
     E_{U}=((\mu+\sigma\sqrt{\frac{2}{\pi}})-(\mu+\sigma\sqrt{\frac{2}{\pi}}))^{2}+\sigma^{2}(1-\frac{2}{\pi})=\sigma^{2}(1-\frac{2}{\pi})
\end{equation}
Equation \ref{eq:EP} and can be calculated as follows:\\
\text{\quad\quad}Since the $i$-th sample is selected with the exponential probability in Equation \ref{eq:P}
\begin{equation}
    E_{P}[A]=\mu-\Lambda\sigma^2+\lozenge
\end{equation}
\begin{equation}
    E_{P}[A^2]=\mu^2+\sigma^2+\Lambda^{2}\sigma^4-2\Lambda\mu\sigma^2+\lozenge(\Lambda\mu^2-2\mu)
\end{equation}
\begin{equation}
     \lozenge=\frac{\sqrt{2}\sigma e^{\frac{-\sigma^2\Lambda^2}{2}}}{\sqrt{\pi}\textit{erfc}(\frac{\sqrt{2}}{2}\sigma\Lambda)}
\end{equation}
\text{\quad\quad}Thus
\begin{equation} 
\begin{aligned}
     E_{P}=((\mu-\Lambda\sigma^2+\lozenge)-\mu)^{2}&+\mu^2+\sigma^2+\Lambda^{2}\sigma^4-2\Lambda\mu\sigma^2+\lozenge(\Lambda\mu^2-2\mu)-(\mu-\Lambda\sigma^2+\lozenge)^2\\
     &=\sigma^{2}(\frac{2}{\pi}+1)+(2\sigma\frac{2}{\pi}+\Lambda\sigma^{2})(\Lambda\sigma^{2}-\lozenge)
\end{aligned}
\end{equation}
Finally, we compare $E_{U}$ and $E_{P}$, as $\lozenge$ contains the erfc function, we evaluate these expressions numerically instead of using analytic equivalent. If we want to get $E_{U}<E_{P}$, we should constrain $\sigma<\frac{\pi}{\Lambda}$ by solving equation
\begin{equation}
     \sigma^{2}(\frac{2}{\pi}+1)+(2\sigma\frac{2}{\pi}+\Lambda\sigma^{2})(\Lambda\sigma^{2}-\lozenge)<\sigma^{2}(1-\frac{2}{\pi})
\end{equation}
Finally 
\begin{equation}\nonumber
     \text{If }\sigma<\frac{\pi}{\Lambda} \text{, Then }E_{U}<E_{P}
\end{equation}
\end{proof}

In \mname, the sample is selected according to its loss after modulating the amplitude $l\rightarrow\kappa(l-\epsilon)$. When it corresponds to the probability density function of the exponential probability in Equation \ref{eq:P}, i.e.
\begin{equation}
    \Lambda e^{-\Lambda l}=\kappa(l-\epsilon)
\end{equation}
To solve this equation, we set $x=\Lambda,a=l,b=\kappa(l-\epsilon)$, then, we can get
\begin{equation}
    xe^{-ax}=b
\end{equation}
Then we change it to 
\begin{equation}
    -axe^{-ax}=-ab
\end{equation}
Using, Lambert W function, we get
\begin{equation}
    x=-\frac{W(-ab)}{a}
\end{equation}
Then
\begin{equation}
   \Lambda=-\frac{W(-\kappa l(l-\epsilon))}{l}
\end{equation}
From Equation \ref{eq:ccl} and Figure \ref{fig:property} (a), when $\lambda<0.01$,\\
for the loss of easy samples ($l-\epsilon< 0$)
\begin{equation}
   -\kappa l(l-\epsilon)>0,\ \text{then } \Lambda < 0 <\frac{\pi}{\sigma}
\end{equation}
for the loss of hard samples ($l-\epsilon\leq 0$)
\begin{equation}
    \kappa(l-\epsilon)\approx 0,\ \text{then } \Lambda \approx 0 <\frac{\pi}{\sigma}
\end{equation}

Thus, when we set $\lambda<0.01$ in the constraint function $\varphi_{C}$, we can satisfy the condition $\sigma<\frac{\pi}{\Lambda}$ in Theorem \ref{theo:2}.
Finally, \mname satisfies Theorem \ref{theo:1} and \ref{theo:2} $\Longrightarrow$ Theorem \ref{theo:3}.

\begin{table}[!ht]\small
\centering
\caption{Notations Used in Appendix}
\setlength{\tabcolsep}{10mm}{
\begin{tabular}{ll}
\toprule[0.8pt]
Notation  &Description \\
\midrule[0.8pt]
$X_{i},x_{i,t}, Y_{i}$ & Time series sample, its value at time $t$, and its label\\
$\mathcal{S},\mathcal{D}$ & Dataset and its distribution \\
$f,w$ & Deep learning model and its parameters\\
\midrule[0.4pt]
$\mathcal{L}$ & Overall loss of a task\\
$l_{i}$ & Original loss when learning sample $X_{i}$\\
$L_{i}$ & New loss of $X_{i}$ after using \mname\\
$\kappa$ & Confidence coefficient to modulate $l$\\
$\lambda$ &  Regularization coefficient to constrain $\kappa$\\
$\epsilon$ &  Threshold to distinguish difficulty of $l$\\
$\varepsilon$ & Indication of current training status\\
$\varphi_{C}$ &Constraint function about $\kappa$\\
$\varphi_{D}$ &Differentiation function about $\epsilon$\\
$\varphi_{S}$ &State function about $\varepsilon$\\
$\mathcal{N}(\mu_{l},\sigma^{2}_{l})$ & Distribution of loss $l$ (mean and variance)\\
$W(*)$ & Lambert W function for the solution to $y=xe^{x}$\\
$\mathcal{F}$ &Periodic function in \mnamee\\
\bottomrule[0.8pt]
\end{tabular}}
\end{table}

\end{document}